\documentclass[acmsmall]{acmart}

\AtBeginDocument{%
  }

\setcopyright{cc}
\setcctype{by}
\acmJournal{PACMHCI}
\acmYear{2025} \acmVolume{9} \acmNumber{2} \acmArticle{CSCW010} \acmMonth{4} \acmPrice{}\acmDOI{10.1145/3710908}

\usepackage{makecell} 
\usepackage{hyperref}
\usepackage{multirow}
\usepackage{longtable}
\usepackage{subcaption}
\usepackage{xcolor}
\usepackage{framed}
\usepackage{soul}

\usepackage{color, soul}
\usepackage{float}
\usepackage{graphicx}

\definecolor{mycolor}{RGB}{0,0,255}
\definecolor{bgcolor}{rgb}{0.9,0.9,1} 
\definecolor{revisionblue}{RGB}{20, 90, 230}

\newif\ifdraft
 \drafttrue
\ifdraft
  \newcommand{\todo}[1]{\textsf{\textbf{\textcolor{red}{[TODO: #1]}}}}
\else
  \newcommand{\todo}[1]{}
\fi

\newcommand\revision[1]{#1}

\begin{document}

\title[EARN Fairness]{EARN Fairness: Explaining, Asking, Reviewing, and Negotiating Artificial Intelligence Fairness Metrics Among Stakeholders}

\author{Lin Luo}
\authornote{Corresponding Author}
\orcid{https://orcid.org/0000-0002-0310-3158}
\email{l.luo.1@research.gla.ac.uk}
\affiliation{%
  \institution{School of Computing Science, University of Glasgow}
  \country{Glasgow, United Kingdom}
}

\author{Yuri Nakao}
\orcid{https://orcid.org/0000-0002-6813-9952}
\email{nakao.yuri@fujitsu.com}
\affiliation{%
  \institution{FUJITSU LIMITED}
  \country{Tokyo, Japan}
}

\affiliation{%
  \institution{Graduate School of Arts and Sciences, The University of Tokyo}
  \country{Tokyo, Japan}
}

\author{Mathieu Chollet}
\orcid{https://orcid.org/0000-0001-9858-6844}
\email{mathieu.chollet@glasgow.ac.uk}
\affiliation{%
  \institution{School of Computing Science, University of Glasgow}
  \country{Glasgow, United Kingdom}
}

\author{Hiroya Inakoshi}
\orcid{https://orcid.org/0000-0003-4405-8952}
\email{inakoshi.hiroya@fujitsu.com}
\affiliation{%
  \institution{Artificial Intelligence Laboratory, FUJITSU LIMITED}
  \country{Tokyo, Japan}
}

\author{Simone Stumpf}
\orcid{https://orcid.org/0000-0001-6482-1973}
\email{simone.stumpf@glasgow.ac.uk}
\affiliation{%
  \institution{School of Computing Science, University of Glasgow}
  \country{Glasgow, United Kingdom}
}

\renewcommand{\shortauthors}{Lin Luo, Yuri Nakao, Mathieu Chollet, Hiroya Inakoshi, and Simone Stumpf}

\begin{abstract}
  Numerous fairness metrics have been proposed and employed by artificial intelligence (AI) experts to quantitatively measure bias and define fairness in AI models. Recognizing the need to accommodate stakeholders' diverse fairness understandings, efforts are underway to solicit their input. However, conveying AI fairness metrics to stakeholders without AI expertise, capturing their personal preferences, and seeking a collective consensus remain challenging and underexplored. To bridge this gap, we propose a new framework, EARN (\textit{Explain, Ask, Review, and Negotiate}) Fairness, which facilitates collective metric decisions among stakeholders without requiring AI expertise. The framework features an adaptable interactive system and a stakeholder-centered EARN Fairness process to \textit{Explain} fairness metrics, \textit{Ask} stakeholders' personal metric preferences, \textit{Review} metrics collectively, and \textit{Negotiate} a consensus on metric selection. To gather empirical results, we applied the framework to a credit rating scenario and conducted a user study involving 18 decision subjects without AI knowledge. We elicited their personal metric preferences and subsequently we studied how they reached metric consensus in team sessions. Our work shows that the EARN Fairness framework supports stakeholders to express and negotiate fairness preferences, and we provide practical guidance for implementing human-centered AI fairness in high-risk contexts. Through this approach, we aim to reach consensus of fairness perspectives, fostering more equitable and inclusive AI fairness.
\end{abstract}

\ccsdesc[500]{Human-centered computing~Empirical studies in HCI}
\ccsdesc[300]{Human-centered computing~HCI design and evaluation methods}
\ccsdesc[500]{Computing methodologies~Artificial intelligence}

\keywords{AI fairness, fairness metrics, human-in-the-loop fairness, stakeholders, fairness consensus}

\received{July 2024}
\received[revised]{October 2024}
\received[accepted]{December 2024}

\raggedbottom

\maketitle
\section{Introduction}
Ensuring artificial intelligence (AI) fairness is crucial for ethical integrity, legal compliance, and human trust \cite{landers_auditing_2023}, particularly in high-stakes scenarios such as healthcare \cite{10.1145/3531146.3533154}, finance \cite{10.1145/3490354.3494408}, and university admission \cite{10.1145/3351095.3372867}. AI fairness experts have made significant efforts to define fairness quantitatively through fairness metrics \cite{10.1145/3457607, 10.1145/3194770.3194776}. These metrics help assess the fairness of AI systems, audit their compliance with specific fairness requirements, and guide unfairness mitigation \cite{hort2023bias}. Currently, there are over 20 popular fairness metrics, categorized into three primary categories — group fairness, subgroup fairness, and individual fairness — to measure whether AI systems treat groups, subgroups, and individuals fairly, respectively \cite{10.1145/3194770.3194776, 10.1145/3457607, 10.1145/3616865}. 

However, there are three primary issues when applying fairness metrics to assess AI fairness. First, choosing the right metrics becomes problematic due to the lack of clear, unified guidelines for metric selection, especially in the absence of a "one-size-fits-all" metric and the insufficiency of using a single metric \cite{10.1145/3457607,10.1145/3512899,castelnovo_clarification_2022, 2024crowdsource/10.1145/3640543.3645209}. Second, AI fairness metric selection, often led by AI experts lacking stakeholder input, risks overly depending on these experts' perspectives and disregards fairness as a societal value with diverse interpretations among stakeholders \cite{10.1145/3544548.3581227, 10.1145/3194770.3194776, 10.1145/3457607, hort2023bias}. This could lead to AI models deemed technically "fair" by AI experts but not accepted in practice because they fail to reflect stakeholders' views on fairness \cite{whittaker2018ai,10.1145/3334480.3375158}. \revision{Previous works \cite{wong2020democratizing, 10.1145/3411764.3445308,10.1145/3375627.3375862,10.1145/3292500.3330664,10.1145/3306618.3314248} have begun to identify stakeholders' metric preferences, for example, Cheng et al. \cite{10.1145/3411764.3445308} developed an interface with an interview protocol to elicit each stakeholder's fairness notions in child maltreatment risk management. However, the range of solicited metrics remains limited, and more work is needed to make fairness metrics more understandable to stakeholders without AI expertise, ensuring fairness is accessible to a broader audience. Last, even when personal fairness perspectives are collected from stakeholders, previous work has shown that different stakeholders often have different fairness goals} \cite{10.1145/3512899, 10.1145/3411764.3445308, doi:10.1080/10447318.2022.2067936,10.1145/3025453.3025884,2024crowdsource/10.1145/3640543.3645209}. \revision{Although there is a growing recognition of the importance of negotiating views on fairness and establishing consensus \cite{10.1145/3544548.3581527, 10.1145/3359284,10.1145/3359283}, there are knowledge gaps about when stakeholders' fairness metric preferences collide, whether consensus can be reached on metric selection, what kind of consensus is reached, and how it is reached. Given these issues, it becomes crucial to not only capture the diverse personal preferences of stakeholders but also investigate fair approaches for negotiating differences and reaching a consensus between multiple stakeholders, ensuring AI fairness aligns with societal values. }

To address the aforementioned \revision{issues} and fill the knowledge gaps, we propose the EARN (\textit{Explain, Ask, Review, Negotiate}) Fairness framework to \revision{elicit personal preferences and support collective negotiation}. Our framework includes two components: an adaptable interactive system, the Fairness Explainer and Explorer (FEE), \revision{which explains six group and subgroup fairness metrics and two individual fairness metrics to stakeholders}, and a stakeholder-centered EARN Fairness process. Stakeholders engage with FEE through the two-phase EARN Fairness process: an individual session to \textit{Explain} fairness metrics to them and \textit{Ask} for their personal preferences on metrics and reasons, followed by a team session to collectively \textit{Review} metric definitions and \textit{Negotiate} a consensus on fairness metrics. 

To ground our investigations in a practical context, we applied the EARN Fairness framework in a credit rating model and recruited 18 decision subjects without AI backgrounds. Through individual sessions, we \revision{elicited stakeholders' personal preferences for fairness metrics. Further, using team sessions, we investigated how they established a collective consensus on fairness metrics among them.} Through this user study, we address the following research questions:
\begin{itemize}
    \item RQ1: \revision{What are stakeholders’ personal preferences on AI fairness metrics that require negotiation?}
    \item RQ2: How do stakeholders negotiate to resolve personal differences in metric preferences? 
\end{itemize}

Our contributions are as follows:
\begin{itemize}
\item \revision{We provide a framework for AI practitioners and researchers to engage stakeholders in reaching a collective agreement on metric selection, without requiring AI knowledge.}
\item We identify strategies that stakeholders employ to reach consensus, which could be used as a blueprint for scaling up the representation of stakeholder views. 
\item \revision{We put forward design recommendations for user interfaces (UIs) to support the ability of stakeholders with no AI knowledge to explore and negotiate AI fairness.}
\end{itemize}

This paper is structured as follows. Section~\hyperref[sec: Related Work]{2} reviews related work on AI fairness metrics, on soliciting stakeholders' metric preferences, and highlights the research gaps. Next, we introduce the EARN Fairness framework in Section~\hyperref[sec: Framework]{3}. Section~\hyperref[sec: User Study]{4} describes a user study of applying the framework in a credit rating scenario. We present the results of our user study, answering each research question in Section~\hyperref[sec: Result]{5}. \revision{Finally, in Section~\hyperref[sec: Discussion]{6}, we discuss limitations, future research, as well as wider implications of our research.}

\section{Related Work}
\label{sec: Related Work}
First, we explore how AI fairness has been conceptualized and measured across different metrics. We then review existing approaches to solicit stakeholders' fairness metrics preferences, and we conclude this section by identifying research gaps. 

\subsection{AI Fairness Metrics}
\label{sec:metrics}
Researchers have proposed various metrics to quantify AI fairness, focusing on three main categories: group fairness and subgroup fairness, which address fairness at the group level, and individual fairness, focusing on fairness at the individual level \cite{10.1145/3194770.3194776, 10.1145/3457607}. Each category has its distinct advantages and limitations. 

Group fairness prioritizes equal treatment among groups, usually defined by \textit{protected} characteristics enshrined by legislation, such as age or gender, or \textit{sensitive} features based on contexts, e.g., foreign worker status. Group fairness is easily comprehensible to policymakers and the public, and it is also feasible to implement computationally \cite{10.1145/3292500.3330664,10.1145/3351095.3372864}. Some prominent group fairness metrics include \cite{10.1145/3457607, hort2023bias}: 
\begin{itemize}
    \item Demographic Parity \cite{NIPS2017_a486cd07} (also known as Statistical Parity \cite{10.1145/2090236.2090255}): Both protected and unprotected group members have the same probability of being predicted as a positive class.
    \item Equal Opportunity \cite{10.5555/3157382.3157469}: Both protected and unprotected groups have the same probability that a subject in the positive class is predicted as a positive class.
    \item Predictive Equality \cite{10.1145/3097983.3098095}: Both protected and unprotected groups have the same probability that a subject in the negative class is predicted as a positive class.
    \item Equalized Odds \cite{10.5555/3157382.3157469}: Simultaneously satisfy both Equal Opportunity and Predictive Equality.
    \item Outcome Test \cite{articleoutcometest}: Both protected and unprotected groups have the same probability that a subject with a positive prediction genuinely belongs to the positive class.
    \item Conditional Statistical Parity \cite{10.1145/3097983.3098095}: Both protected and unprotected groups are equally likely to be predicted as a positive class, under conditions controlled by a set of \textit{legitimate} features L. \textit{Legitimate} features are those directly relevant and considered appropriate for influencing decision-making outcomes. For instance, in a credit rating scenario, legitimate features are factors that affect an applicant's creditworthiness, such as the applicant’s credit history and employment status \cite{10.1145/3194770.3194776}. 
\end{itemize} 

However, group fairness and its metrics have been criticized for potentially overlooking intersectional bias caused by the combination of multiple protected features, e.g. black women. Thus researchers have introduced subgroup fairness, applying group fairness metrics on multiple protected features to ensure fairness over subgroups \cite{9101635,Kobayashi2022One}. Challenges include dealing with the potentially vast number of feature combinations and addressing issues like data sparsity, where some subgroups have very few or no observations, making it difficult to assess fairness accurately \cite{pmlr-v80-kearns18a, 10.1145/3287560.3287592, castelnovo_clarification_2022}. 

Fairness at the group level can still mask individual injustices \cite{10.1145/2090236.2090255,10.1145/3351095.3372864}. Therefore, individual fairness is also gaining attention. Popular metrics include:

\begin{itemize}
    \item Consistency \cite{pmlr-v28-zemel13}: Similar individuals will be treated similarly.
    \item Counterfactual Fairness \cite{NIPS2017_a486cd07}: Individuals receive the same results across both the actual world and a counterfactual world in which the individual belongs to a different demographic group.
\end{itemize} 

Despite being theoretically attractive, individual fairness also poses challenges in practical application, notably in calculating similarity, which involves defining similarity and determining the scope of individuals to be considered in these comparisons \cite{10.1145/3351095.3372864, pmlr-v28-zemel13, 8843908}.

Metrics such as these have been integrated into tools to enable AI experts to assess fairness, such as FairSight \cite{8805420}, Silva \cite{10.1145/3313831.3376447}, FairVis \cite{8986948}, and FairRankVis \cite{9552229}. Additionally, open-source AI fairness toolkits like IBM's AI Fairness 360 \cite{8843908}, Google's What-If \cite{8807255}, Microsoft's Fairlearn \cite{weerts2023fairlearn}, and DALEX \cite{10.5555/3546258.3546472} offer comprehensive approaches for analyzing and mitigating unfairness in AI systems, built around these metrics.

Even once metrics are established, it is still to be determined at what level this metric reaches fairness, i.e., fairness thresholds. For example, the Demographic Parity Ratio (DPR) of 0.8 serves as a pivotal fairness threshold according to US law \cite{10.1145/2783258.2783311}; if a system meets or exceeds this value, it is considered "fair" (enough). While these fairness thresholds are crucial, there is also a lack of universal standards for them \cite{saleiro2019aequitas}. 

\subsection{Soliciting Stakeholders' Fairness Metric Preferences}
AI fairness experts need to select from numerous metrics in a specific application context, yet there is no agreement on the most appropriate metrics in specific contexts \cite{hort2023bias,challengesurvey}. This has motivated numerous studies to solicit \revision{stakeholders' personal metric preferences} \cite{wong2020democratizing,10.1145/3411764.3445308, 10.1145/3306618.3314248,2024crowdsource/10.1145/3640543.3645209,ilvento:LIPIcs.FORC.2020.2}.

\revision{Explaining metrics to stakeholders without any AI background is challenging due to the complex mathematical concepts of these metrics \cite{10.5555/3524938.3525714}}. Many studies opt to design various forms of user input to indirectly map users' metric preferences. One popular method is model selection, where stakeholders' metric preferences are inferred from their chosen models \cite{10.1145/3292500.3330664,2024crowdsource/10.1145/3640543.3645209,10.1145/3351095.3372831}. For example, laypeople have been asked to identify which of two models is more discriminatory based on their prediction labels \cite{10.1145/3292500.3330664}, or to compare three hypothetical ML models that adhere to different fairness metrics to determine the preferred one \cite{2024crowdsource/10.1145/3640543.3645209}. There are also other forms of input designed to indirectly solicit stakeholders' metric preferences \cite{10.1145/3375627.3375862,10.1145/3306618.3314248,umapworkshop}. For example, in loan scenarios, researchers collected stakeholders' fund allocation decisions to match their preferred metrics \cite{10.1145/3306618.3314248}, or allowed lay users to label AI outcomes as "fair" or "unfair" and to adjust feature weights used for model training, thereby calculating which metric has been optimized to uncover their preferences \cite{umapworkshop}. These indirect solicitation methods reduce people's cognitive demands, yet mask the underlying metrics, potentially resulting in choices that inadequately map stakeholders' true fairness perspectives. Moreover, the user input formats and metrics, along with their application context, are often highly interrelated and need careful design and alignment, thereby lacking the flexibility to adapt to different scenarios.

To avoid these, some approaches aim to directly explain fairness metrics to non-AI experts through visualizations. Previous work \cite{10.1145/3411764.3445308} has shown the value of communicating group fairness metrics in a simplified and concrete form that allows metric exploration and preference elicitation. Other work has developed a range of UI components to assist stakeholders, e.g., domain experts \cite{10.1145/3411764.3445308, doi:10.1080/10447318.2022.2067936} and end users \cite{IUI2021DesignMF,10.1145/3514258} in exploring fairness within datasets and AI models. However, due to the complexity of the concepts behind these metrics, existing works rely on simplified textual and graphical explanations, lacking interactive designs that would allow users to delve into the details of metrics \cite{10.1145/3411764.3445308, 10.1145/3306618.3314248, 10.5555/3524938.3525714}. This may hinder stakeholders' understanding of different fairness metrics, thereby limiting their ability to provide insights or critiques on metric selection.

\revision{Once personal preferences for measuring fairness have been elicited, there is still a need to manage the differences in personal preferences.} For instance, in loan scenarios, data scientists and domain experts have different perspectives on fairness \cite{doi:10.1080/10447318.2022.2067936}, with data scientists preferring group fairness and domain experts focusing mainly on individual fairness. Even within the same stakeholder types, people hold different metric preferences, for example, in child maltreatment risk management scenarios, some prefer equalized odds, while some prefer statistical parity \cite{10.1145/3411764.3445308}. Hence, balancing diverse fairness needs and seeking consensus among stakeholders is crucial for implementing fair AI successfully. Current research primarily focuses on seeking consensus on AI model building or predictive outcomes through participatory design \cite{10.1145/3579601, 10.1145/3359283} or co-design \cite{IUI2021DesignMF,10.1145/3514258} for AI fairness democratization. For example, in the context of goods division, Lee et al. \cite{10.1145/3359284} reached distribution outcome consensus by allowing individual adjustments first, then group discussions for collective decision-making on the allocations. Furthermore, Lee et al. \cite{10.1145/3359283} introduced the "WeBuildAI" Framework to achieve a consensus-based algorithm driven by voting for fair food donation. \revision{While recent research advocates and explores negotiating different fairness notions among individuals, it has not translated into practical, quantifiable fairness metrics \cite{10.1145/3544548.3581527}. To the best of our knowledge, the process of stakeholders negotiating the consensus on fairness metric selection has not been explicitly explored}.

\subsection{Research Gaps}
In summary, the current research has \revision{two clear gaps. First, indirect methods used to elicit metric preferences might reduce transparency and hinder stakeholders' understanding of AI fairness. More detailed explanation work, like that conducted by Cheng et al. \cite{10.1145/3411764.3445308}, which directly explains metrics, is necessary and should be prioritized to make fairness more comprehensible to a wider audience. Second, there is a lack of research on how to address differences and achieve consensus between individuals once personal preferences have been elicited.}  

Our work addresses these gaps by designing an adaptive interactive system that visually communicates three fairness categories, including eight metrics, to stakeholders without requiring prior AI knowledge. \revision{Through a novel process in which this interactive system is used, we aim to elicit stakeholders' personal preferences and then help to reconcile differences to reach consensus on metrics to apply.}

\section{The EARN Fairness Framework}
\label{sec: Framework} 
We present the EARN (\textit{Explain, Ask, Review, Negotiate}) Fairness framework to uncover a collective metric selection from stakeholders without necessitating prior AI expertise. This framework consists of two components: a stakeholder-centered EARN Fairness process and an adaptable interactive system, the Fairness Explainer and Explorer (FEE). The stakeholder-centered EARN Fairness process comprises two phases: (1) individual sessions to \textit{Explain} fairness metrics and \textit{Ask} for stakeholders' personal preferences on metrics and reasons; followed by (2) team sessions to collectively \textit{Review} metric definitions and \textit{Negotiate} a consensus on fairness metrics. FEE supports both phases of the EARN Fairness process, allowing the exploration of fairness metrics.

\subsection{The EARN Fairness Process}
\label{sec: process}
\subsubsection{Individual Session to Explain and Ask}
\label{sec: Individual Session Method}
Our framework begins with individual sessions, which are designed to discover each stakeholder's personal preferences, with one session per stakeholder. The \textit{Explain} part is woven throughout the entire session, where each stakeholder interacts with the Fairness Explainer and Explorer (FEE) to delve into fairness metrics and model fairness. The \textit{Ask} part requires stakeholders to reflect on three questions:
\begin{itemize}
    \item \revision{Question 1: What are the three most preferred metrics, and in what order?}
    \item \revision{Question 2: What are the fairness thresholds for different fairness categories?}
    \item \revision{Question 3: What are the reasons behind their choices?}
\end{itemize}

\revision{These questions are used to elicit personal preferences and reasons for these choices to support later negotiation.}
\subsubsection{Team Session to Review and Negotiate}
\label{sec: Group Session Method}
Our framework then shifts to a collaborative team session, \revision{where personal metric preferences are discussed and negotiated into a team consensus among stakeholders}. This session involves two key tasks: team members first collectively \textit{Review} fairness metric definitions using FEE to establish a shared understanding, and then \textit{Negotiate}\revision{, where they can freely utilize FEE to compare individual preferences and highlight any potential areas of disagreement}. The team needs to accomplish \revision{selecting or defining a fairness metric that aligns with or reconciles all team members' fairness preferences.} 

\subsection{The Fairness Explainer and Explorer (FEE)}
\label{sec:Fairness Explainer and Explorer UI}
We designed the Fairness Explainer and Explorer (FEE) to support the EARN Fairness process. The FEE is an adaptive interactive system that explains AI fairness metrics, enabling users to explore, understand, and identify their preferred metrics. We will detail its design goals, supported metrics, and UI components, and provide a comparison of FEE with previous fairness UIs.

\subsubsection{Design Goals}
Our target users are individuals with no background in AI. To help users comprehend and compare different metrics, we draw inspiration from prior research \cite{doi:10.1080/10447318.2022.2067936, 10.1145/3411764.3445308, IUI2021DesignMF}, and propose four design goals: a.) \textbf{Contextual Understanding:} help users develop a contextual understanding of the model by analyzing AI model predictions case-by-case, b.) \textbf{Model Transparency:} explaining basic AI concepts, offering model explanations and model performance information to enhance users’ comprehension of the AI model, c.) \textbf{Diverse Explanation and Comparison:} elucidating metrics in diverse styles and plain language, indicating when a metric shows the model to be unfair, presenting fairness metric results at the case level and providing intuitive comparisons across metrics, and d.) \textbf{Fairness Adjustability:} allowing users to question biases in original data and AI model predictions, and dynamically tracking their impact on model fairness. Moreover, our UI design and metric explanation methods are fundamentally rooted in the underlying mathematical logic of the metrics, allowing them to be adaptive and applicable across different domains.

\subsubsection{Fairness Metrics Covered in FEE}
We have expanded the range of fairness metrics beyond prior work in both metric quantity and fairness category \cite{wong2020democratizing, 10.1145/3411764.3445308,10.1145/3375627.3375862,10.1145/3292500.3330664,10.1145/3306618.3314248}, introducing eight metrics covering group, subgroup, and individual fairness. Offering a broader range of fairness metrics helps meet the varied expectations and needs of stakeholders and also provides AI experts with more diverse insights into metric selection. Further, given that broader stakeholders, especially those without AI backgrounds, might have divergent understandings of fairness compared to AI experts, our fairness metrics include both widely used metrics in AI fairness practice and relatively less popular ones \cite{10.1145/3457607, 10.1145/3194770.3194776, hort2023bias}. See Section~\hyperref[sec:metrics]{2.1} for definitions and Appendix~\ref{app:Fairness Metrics and Explanations} for how we implemented these fairness metrics. All metric results are presented as percentages.

For group fairness, we provide five prevalent metrics in AI fairness practice: Demographic Parity \cite{10.1145/2090236.2090255, NIPS2017_a486cd07}, Equal Opportunity \cite{10.5555/3157382.3157469}, Equalized Odds \cite{10.5555/3157382.3157469}, Predictive Equality \cite{10.1145/3097983.3098095}, and Outcome Test \cite{articleoutcometest}. We also include one less common metric, Conditional Statistical Parity \cite{10.1145/3097983.3098095}. Moreover, we apply these metrics to subgroups to support subgroup fairness exploration. 
Consistent with prior research targeting laypeople \cite{2024crowdsource/10.1145/3640543.3645209,10.1145/3411764.3445308}, we represent group and subgroup fairness metric results using the maximum difference between (sub)groups.

For individual fairness, we used two popular individual fairness metrics: Counterfactual Fairness \cite{NIPS2017_a486cd07} and Consistency \cite{pmlr-v28-zemel13}. Counterfactual fairness is quantified by the proportion of instances where decisions remain consistent between the actual and counterfactual worlds. The Consistency score is computed based on the similarity of AI predictions among the five nearest neighbors, as per IBM's AI Fairness 360 \cite{8843908}. 

Our framework also allows fairness threshold settings.  Group-level metric results range from 0\% to 100\%, indicating increasing disparities among (sub)groups. Therefore, a threshold of 0\% means absolute fairness and higher values signify higher unfairness tolerance. In contrast, individual fairness metric results range from 0\% to 100\%, indicating the degree of fairness in treating individuals. Thus, a threshold of 100\% means absolute fairness and lower values signify higher unfairness tolerance. 

\begin{figure}[h]
\centering
\includegraphics[width=0.85\linewidth]{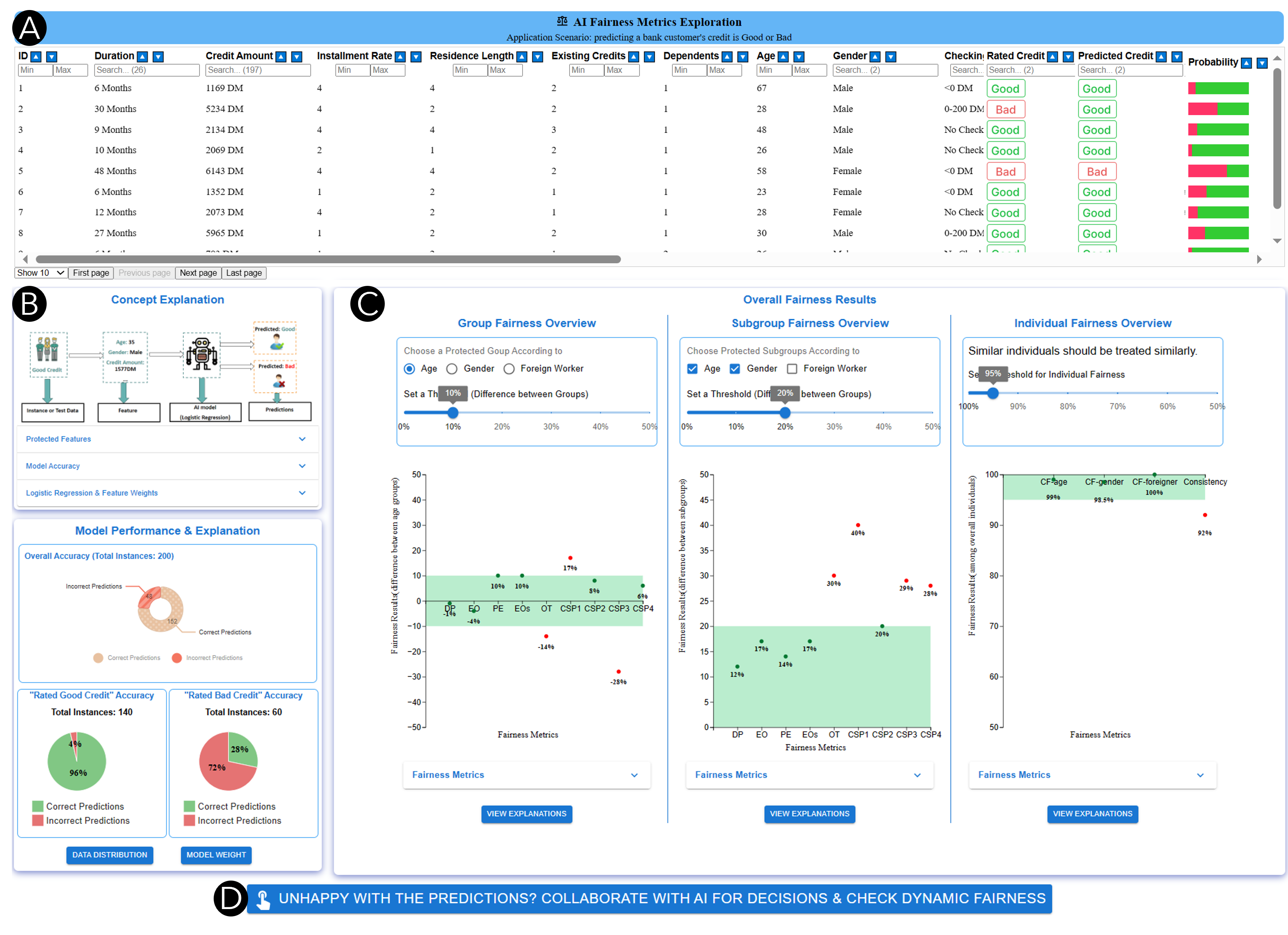}
\caption{Fairness Explainer and Explorer: UI Dashboard. }
\label{fig: UI_Dashboard}
\end{figure}

\begin{figure}[h]
\centering
\includegraphics[width=0.85\linewidth]{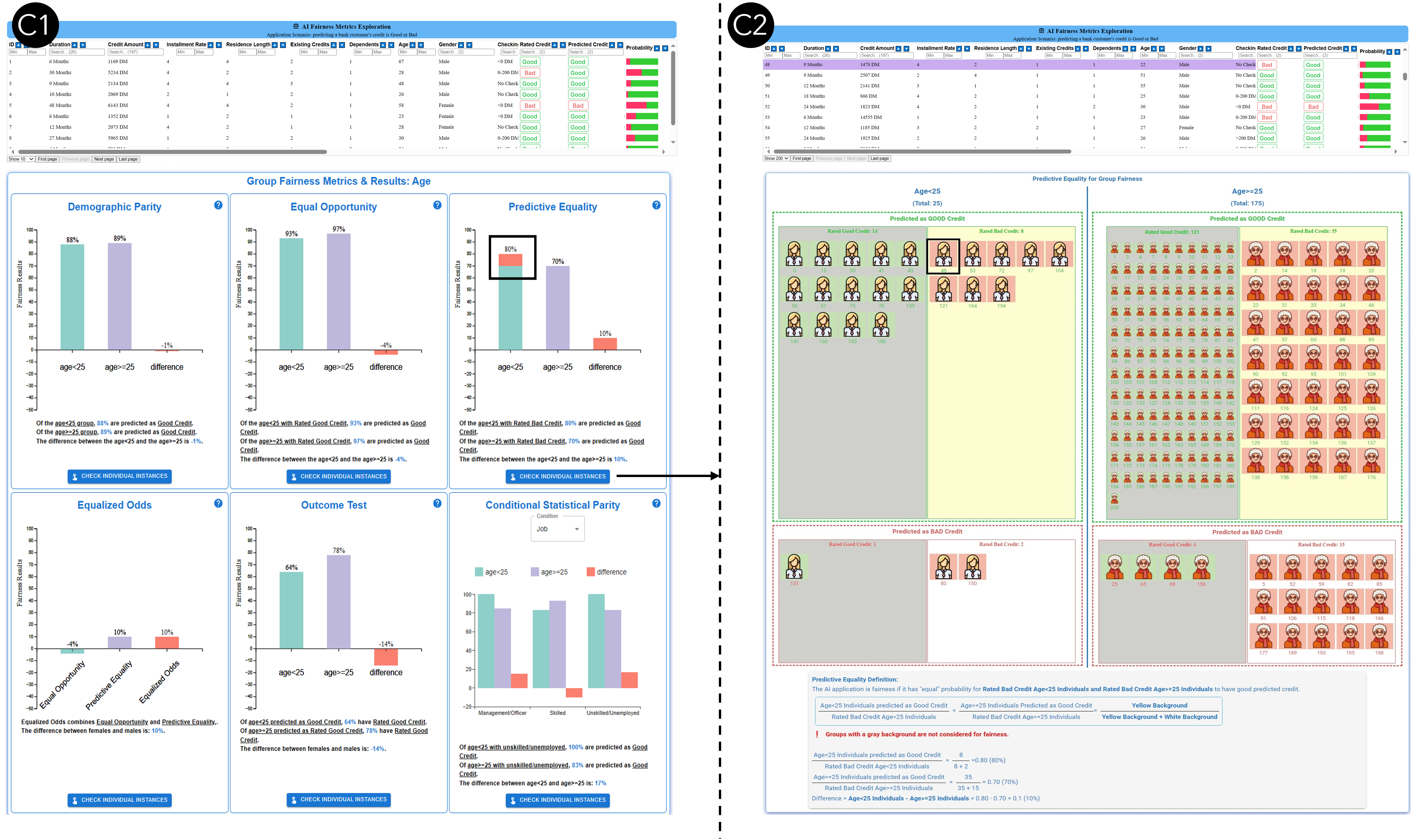}
\caption{Static Fairness Exploration Component (Explanation Views for Group Fairness). This component is activated when users click "VIEW EXPLANATIONS" in the group fairness panel of the UI dashboard (Figure~\ref{fig: UI_Dashboard}), showing detailed explanatory views for each group fairness metric. C1 shows a user selecting the "Age" protected feature to view explanations for all group fairness metrics. C2 depicts a user clicking "CHECK INDIVIDUAL INSTANCES" to examine the instance-level explanation for the Predictive Equality metric.}
\label{fig: Group_Explanation}
\end{figure}

\begin{figure}[h]
\centering
\includegraphics[width=0.85\linewidth]{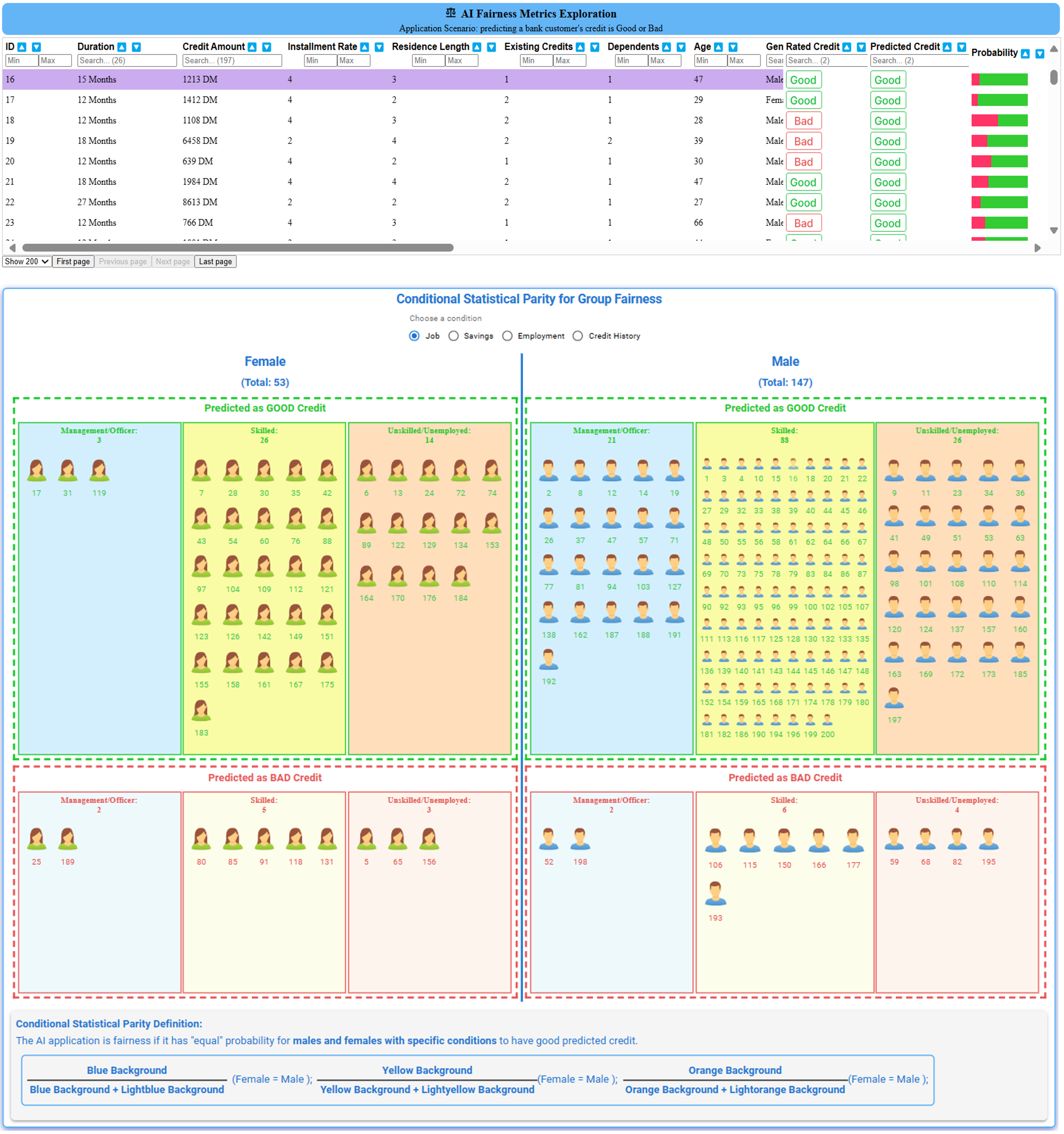}
\caption{This is another example of instance-level explanation views for group fairness metrics related to the protected feature of "Gender". After users set the condition to "Job" and click "CHECK INDIVIDUAL INSTANCES" for Conditional Statistical Parity (in Figure~\ref{fig: Group_Explanation} C1), they will see the corresponding instance-level explanation. Users can switch between conditions using radio buttons, such as "Credit History".}
\label{fig: CSP_Gender_Job_View}
\end{figure}

\begin{figure}[h]
\centering
\includegraphics[width=0.85\linewidth]{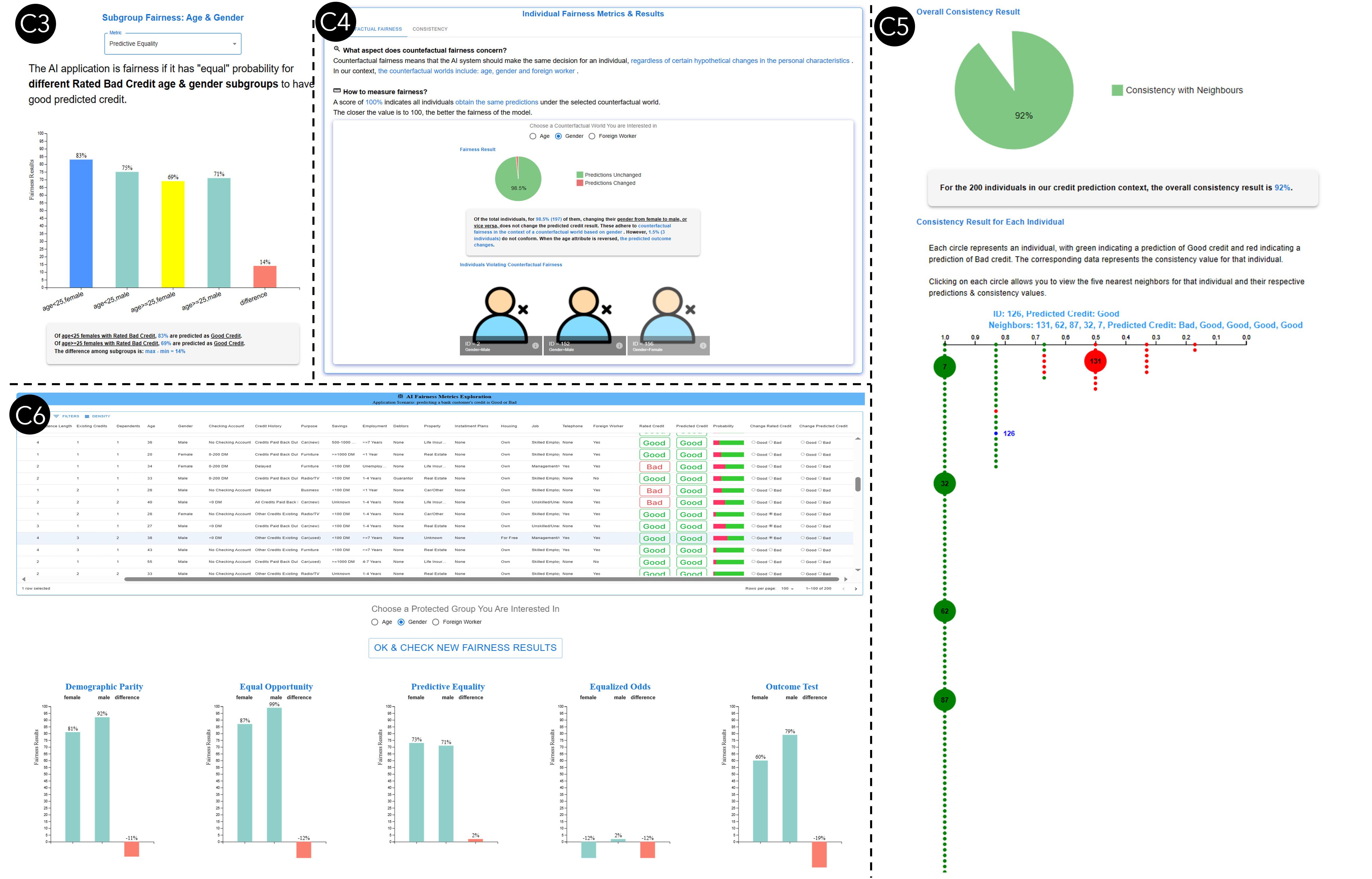}
\caption{Static Fairness Exploration Component (C3:Explanation Views for Subgroup Fairness; C4 and C5:Explanation Views for Individual Fairness) \& Dynamic Fairness Exploration Component (C6).  }
\label{fig: SubgroupIndividual_Explanation}
\end{figure}

\subsubsection{UI Components}
The FEE consists of four main components (Figure~\ref{fig: UI_Dashboard}): Data Exploration (Component A), Model Exploration (Component B), Static Fairness Explanation and Exploration (Component C), and Dynamic Fairness Exploration (Component D), each addressing its respective goal: a), b), c), and d). Detailed UI views for each component are provided in the supplementary material.

\textbf{Data Exploration:} This component allows users to explore datasets and AI model predictions (Figure~\ref{fig: UI_Dashboard} A). Each row represents a test instance, with horizontal scrolling to view all features. The last three columns are fixed, showing ground-truth labels ("Rated Credit" by experts), predictions ("Predicted Credit" by the AI model), and model confidence ("Probability") \cite{doi:10.1080/10447318.2022.2067936}. Model confidence is displayed using a color-coded bar chart, and hovering reveals probability values for various predictions. Users can sort and filter instances by feature values to identify potential biases against certain groups or individuals, such as filtering instances based on gender and prediction results to observe how the model performs for different genders. 

\textbf{Model Exploration:} This component lowers the barrier to understanding AI through visual explanations of model parameters and performance (Figure~\ref{fig: UI_Dashboard} B). In the "Concept Explanation" view, users can check the AI model's workflow for prediction and click for concept explanations like \textit{protected features} and \textit{model accuracy}. In the "Model Performance \& Explanation" view, users see visual explanations of model performance: pie charts depicting overall model accuracy and class-specific accuracy, and bar charts showing distributions of positive and negative predictions for different features' values (or bins for continuous features). By clicking "MODEL WEIGHT", users can inspect feature weights, identifying the importance of each feature and its positive (green bar) or negative (red bar) impact on prediction outcomes. This enables users to grasp which features are pivotal in predictions and assess if these features might lead to unfair outcomes. 

\textbf{Static Fairness Explanation and Exploration:} This component offers interactive visualizations of fairness metrics and assessment results of AI fairness (Figure~\ref{fig: UI_Dashboard} C), including three panels: "Group Fairness Overview", "Subgroup Fairness Overview", and "Individual Fairness Overview". Each panel works in a similar way, initially presenting a comprehensive overview displaying all metrics' results within this fairness category. \revision{For group and subgroup fairness, users can select and switch between different protected features or their combinations to assess fairness. In light of prior work \cite{doi:10.1080/10447318.2022.2067936}, users can also adjust the fairness threshold for each category by dragging a slider, which dynamically alters a green rectangular area representing the "fairness zone". Metrics inside this green area, appearing as green dots, indicate the model is fair at the current threshold. Metrics outside the area are red dots, signaling the model is unfair. During this process, users can intuitively visualize AI fairness assessment results for the current metric, compare fairness outcomes across different metrics and categories to set fairness thresholds, or consult detailed explanations further to refine their threshold settings.}

\revision{By} clicking the "VIEW EXPLANATIONS" button within each panel, users can \revision{next} access detailed explanations for metrics under each fairness category and select their top three preferred metrics. These explanations use text, charts, and images to break down fairness metrics at the individual case level, reducing the need for a mathematical or statistical background. Specifically:
\begin{itemize}
    \item \textbf{Group Fairness:} 
    For instance, when users select "Age" and click on "VIEW EXPLANATIONS" in the "Group Fairness Overview" panel, they are directed to the explanation page (Figure~\ref{fig: Group_Explanation} C1). Here, they can explore each group fairness metric by textual definitions and simple bar graphs displaying detailed information for protected groups ("age < 25"), non-protected groups ("age >= 25"), and the corresponding metric results ("difference"). As users hover over the "difference" bar, the UI dynamically highlights the source of the metric result on the advantaged group's bar, as indicated in C1 by the black box annotation. 
    
    If users struggle to understand a fairness metric, e.g., Predictive Equality, they can click "CHECK INDIVIDUAL INSTANCES" to access instance-level explanations (Figure~\ref{fig: Group_Explanation} C2). In this view, users see an image view with each image representing an instance. Images with a red background indicate a "Rated Credit" of "Bad", while a green background indicates a "Rated Credit" of "Good". We further categorize these instances into different data groups by key feature values involved in the metric, i.e., protected feature values, "Rated Credit" values (ground-truth labels), and "Predicted Credit" values (predicted labels). Users can click on each image to view all information for that instance; the Data Exploration component will automatically locate and highlight the clicked instance in purple. This links metric results with concrete examples, facilitating understanding of the metric application at the instance level. For Conditional Statistical Parity, instances are uniquely categorized by protected features, legitimate features, and predicted labels, based on the metric's formula, as shown in Figure~\ref{fig: CSP_Gender_Job_View}.
    
    Our interactive system also distinguishes different data groups with unique background colors, each labeled with a title and the total number of instances within. We employ these titles, color blocks, and the specific number of instances to illustrate the metrics' formulae, as demonstrated in the gray rectangular area below Figure~\ref{fig: Group_Explanation} C2 and Figure~\ref{fig: CSP_Gender_Job_View}. This visually simplifies complex calculations and enhances user comprehension through intuitive color blocks.

    \item \textbf{Subgroup Fairness:} We apply the same group fairness metrics for subgroup fairness exploration. Users can select any subgroup combinations of protected features in the "Subgroup Fairness Overview" panel, such as, "Age" and "Gender", and click "VIEW EXPLANATIONS" to access the explanation page (Figure~\ref{fig: SubgroupIndividual_Explanation} C3). The Data Exploration component remains at the top of the UI, showing only the subgroup fairness view in Figure~\ref{fig: SubgroupIndividual_Explanation} C3. For each metric, we provide textual definitions and bar graphs detailing subgroups and results. When users hover over the difference bar, it dynamically highlights the most disadvantaged (in yellow) and advantaged (in blue) subgroups.
    
    \item \textbf{Individual Fairness:} When users click on "VIEW EXPLANATIONS" in the "Individual Fairness Overview" panel, they are directed to Figure~\ref{fig: SubgroupIndividual_Explanation} C4. Users can click the tabs to switch between Counterfactual Fairness and Consistency. In the "COUNTERFACTUAL FAIRNESS" view (Figure~\ref{fig: SubgroupIndividual_Explanation} C4), users can see the textual explanations of Counterfactual Fairness. Next, users can select diverse protected features, such as "Gender", to view a pie chart and textual explanations of the metric result. Moreover, users can check instances violating counterfactual fairness and click on each icon to view all information for this instance; the Data Exploration component will automatically locate and highlight the clicked instance in purple. 
    
    In the "Consistency" view (Figure~\ref{fig: SubgroupIndividual_Explanation} C5), a pie chart displays the Consistency results for all test instances. Users can explore the consistency of each instance through a scatter plot, where the x-axis represents consistency values, indicating their fairness level. The green and red dots signify AI predictions of Good Credit or Bad Credit. For example, when users select a dot like ID126, the UI zooms in to display the five nearest neighbors, listing their IDs along with the AI's predictions for each. This function not only visually distinguishes similar instances but also indicates fairness level: if all five neighbors share the same color as the selected instance, it signifies that similar instances receive uniform predictions, marking the prediction for the selected instance as fair.  
\end{itemize}

\textbf{Dynamic Fairness Exploration:} During exploration, users might discover that certain groups or individuals have been unfairly treated. Users can navigate to the view in Figure~\ref{fig: SubgroupIndividual_Explanation} C6 by clicking the button in Figure~\ref{fig: UI_Dashboard} D. By modifying ground-truth labels or AI predictions and then selecting protected features, users can reassess the effects on AI group fairness in real-time. This not only provides users with the opportunity to question the fairness of expert or AI predictions but also helps them gain a deeper understanding of fairness metrics.

\subsubsection{Comparison with Previous Fairness UIs}
Leveraging Nakao et al.'s work \cite{doi:10.1080/10447318.2022.2067936} which provided user requirements of fairness investigation tools in a loan scenario \footnote{We extracted AI fairness exploration requirements from loan officers' perspective, as we believe this group more closely aligns with our target audience, who are stakeholders without AI knowledge.}, \autoref{tab:UI comparison} summarizes a comparison between our UI, FEE, and two other tools, FET \cite{10.1145/3411764.3445308} and FairHIL \cite{doi:10.1080/10447318.2022.2067936}. Our tool, FEE, differs by supporting custom threshold inputs and dynamically explaining when metrics indicate fairness or unfairness. It also enhances the transparency of AI fairness by providing detailed model parameters and instance-level fairness metric explanations.
\begin{table}
  \caption{UI Comparison between FET, FairHIL, and our tool, FEE. }
  
  \label{tab:UI comparison}
  \resizebox{\columnwidth}{!}{%
  
  \begin{tabular}{cclllll}
    \toprule
    Area&Use&Requirement &FET &FairHIL & FEE (our tool)\\
    \midrule
    1. Attribute overviews& Informational& 1.1. Attributes, number of records and attribute value distributions& \checkmark & \checkmark  & \checkmark \\
    & & 1.2. Amount of missing data& & & &\\
    & & 1.3. Fairness metrics for model and individual protected attributes&  \checkmark &  \checkmark & \checkmark\\
    & & 1.4. Target distribution&  &  \checkmark &  \checkmark\\
    2. Investigate relationships between attributes & Informational& 2.1. Distribution of protected attributes with other attributes&  & \checkmark &  \\
    & & 2.2. Distribution of user-selected attribute values (e.g., job) and target values &  &  \checkmark  & \checkmark\\
    & & 2.3. Distribution of two user-selected attributes' values &  &  \checkmark  & \\
    & Functional & 2.4. Support creation of new attributes (i.e., calculated from other attributes) &  & \checkmark  &  \\
    & & 2.5. Ability to create/include own fairness metric (if not already in system) &  & \checkmark  &  \\
    & & 2.6. Allow creation of subgroups based on a combination of attributes and see their distribution on target & \checkmark & \checkmark  & \checkmark \\
    & Adjust model& 2.7. Input custom thresholds to affect AI model& &  & \checkmark \\
    3. Individual cases& Informational & 3.1. Specific application and attribute values& \checkmark & \checkmark  & \checkmark \\
    & & 3.2. Fairness metric for individual case &  &  & \checkmark \\
    & & 3.3. Level of similarity between cases & \checkmark & \checkmark  & \checkmark \\
    & & 3.4. Select specific cases to compare and show which attributes are similar & \checkmark & \checkmark & \checkmark  \\
    & & 3.5. Show decision boundaries &  & \checkmark & \checkmark  \\
    & Functional & 3.6. See “What If” results on target based on changes to attribute values &  &  & \checkmark  \\
    4. Model & Informational & 4.1. How model works &  & \checkmark  & \checkmark  \\
    &  & 4.2. How it was created, rationale for decisions in modeling &  &  & \checkmark  \\
    &  & 4.3.  Who created it &  &  & \checkmark \\
    &  & 4.4.  Explanations of when measures indicated unfairness or discrimination &  &   & \checkmark  \\ 
 \bottomrule
\end{tabular}
}
\end{table}

\section{User Study in a Credit Rating Scenario}
\label{sec: User Study}
We applied the EARN Fairness framework in a credit rating scenario, where 18 stakeholders without a background in AI participated in a user study.

\subsection{Data and Model}
\label{sec:Data and Model}
We applied the German Credit Dataset \cite{dataset} due to its popularity in AI fairness research and providing a familiar high-risk banking scenario. It comprises 1000 instances, each with 20 features (13 categorical and 7 numerical features) for classifying "Good" or "Bad" credit, with no missing data. Our study investigated fairness on three features: age and gender, recognized as protected characteristics under law regulations \footnote{\label{website1}Equality Act 2010: \url{https://www.legislation.gov.uk/ukpga/2010/15/contents} \newline
\label{website2}Charter of Fundamental Rights of the European Union, 2012: \url{https://eur-lex.europa.eu/legal-content/EN/TXT/?uri=CELEX:12012P/TXT}}, and foreign worker status as a sensitive feature utilized by many fairness research \cite{4909197, 10.1145/3287560.3287589}. They were processed as follows 1) Age: we converted age into a binary feature, with age<25 as the protected group; 2) Gender: we extracted this from the "personal status and gender" feature, and selected "female" as the protected group; 3) Foreign Worker: we chose "yes" as the protected (sensitive) group. We provided four non-protected features as possible legitimate features in a credit rating scenario: job, savings, employment, and credit history \cite{10.1145/3194770.3194776} for measuring Conditional Statistical Parity.

We employed a logistic regression classifier. Logistic regression's simplicity and interpretability make it ideal for improving model understanding, particularly for non-experts \cite{10.1016/j.inffus.2019.12.012}. Previous research has also effectively employed logistic regression to study fairness in this dataset \cite{hort2023bias}. We used all 1000 instances, training this model with 5-fold cross-validation to ensure sufficient test instances for subgroup fairness analysis. To balance model accuracy and fairness, our model achieves a 0.76 accuracy on 200 test instances. 

\subsection{Participants}
We recruited 18 participants for the study through public and university social media channels, all of whom are decision subjects impacted by AI decisions and have no AI background. Participants were aged 18 to 47 (mean = 28), with diverse genders, professions, nationalities, and ethnicities (Appendix~\ref{app:Participant Information}). Each participant received a £20 Amazon voucher as compensation. We obtained informed consent from them, and each participant was assigned a unique ID for anonymity throughout the study. This study was reviewed and approved by the University of Glasgow College of Science \& Engineering Ethics Committee (Application Number: 300230024).

\subsection{Procedure}
\label{sec: UserStudyProcedure}
Each participant attended two one-hour sessions: an individual session and a team session, both facilitated by a researcher. During the individual session, each participant first filled out a pre-study questionnaire to gather demographic data and then watched a video tutorial on how to use FEE. They then followed the EARN Fairness framework to complete \textit{Explain} and \textit{Ask}. Throughout this session, participants were encouraged to think aloud their thoughts, including but not limited to their perspectives on fairness metrics, as well as provide feedback on our interactive system design and user experience. 

After these individual sessions with 18 participants, a researcher analyzed their top 3 metric ranking lists and formed three teams, each consisting of six participants. To ensure diversity and thus necessitate negotiations, we employed hierarchical sampling to assign participants with similar preferences into different teams. Each participant's metric preference was mapped to an 8-dimensional vector with each dimension representing a specific metric, enabling us to cluster participants by similar metric choices. The final teams were established as follows: Team 1 (P4, P5, P7, P8, P11, P12), Team 2 (P1, P2, P6, P15, P16, P17), and Team 3 (P3, P9, P10, P13, P14, P18) \footnote{P3 did not attend the team session.}.

Finally, we followed the framework to conduct team sessions for each team to complete \textit{Review} and \textit{Negotiate}. Members were provided with hard copies outlining \revision{their tasks and engaged in free-form negotiation without interference from researchers. Guidance was provided that the negotiation results should be unanimously approved by all team members. During the negotiation process, members were free to use FEE}. Researchers aimed to ensure a harmonious negotiation atmosphere and made preparations to offer alternative collective decision-making methods when needed, such as the Borda voting method due to its simplicity and robustness \cite{10.1145/3359283, pmlr-v97-kahng19a}.

\subsection{Data Collection and Analysis}
We collected participants’ top 3 metric ranking lists and fairness thresholds, along with audio recordings and the researcher’s observation notes. We used descriptive statistics to analyze the top 3 metric ranking lists, reporting the raw frequencies for each metric within the top 3 ranks, and to calculate the mean values and standard deviations (SD) for fairness thresholds. Further, we transcribed approximately 20 hours of recordings from all 18 participants. We conducted a thematic analysis \cite{doi:10.1191/1478088706qp063oa} to generate insights into participants' \revision{diverse} preferences on fairness metrics and thresholds, the reasons behind them, and user feedback \revision{on FEE design}. Our research team met weekly to iterate over the themes and coding, reviewing and validating them. After 6 rounds of iteration, we finalized the final codebook (Appendix~\ref{app:codebook}).

For team sessions, we collected negotiation results on metric preferences along with corresponding video and audio recordings. Approximately 4 hours of audio recordings were transcribed, aided by videos for participant ID identification during transcription. For all three teams' negotiation results, we analyzed each team's metric consensus and conducted a thematic analysis of the transcribed data with 4 rounds of iterations to identify the negotiation strategies employed. 

\section{Results}
\label{sec: Result}
In this section, we tackle each of our research questions in turn by evidencing these from data gathered during our user study. 

\subsection{\revision{What are stakeholders’ personal preferences on AI fairness metrics that require negotiation?} (RQ1)}
\label{sec:IndividualResult}

\subsubsection{Preferences for Fairness Metrics}
\autoref{tab:FairnessPreference} shows all the 18 participants’ preferences for fairness metrics and fairness thresholds. We first investigated preferred fairness metric categories that \revision{demonstrate differences to be negotiated}. Individual fairness emerged as the \revision{slightly more} favored fairness metric category, chosen by 9 participants in their Top-1 metric, closely followed by subgroup fairness, chosen by 7 participants. In contrast, despite group fairness being widely favored in industry and academia for its ease of implementation \cite{10.1145/3351095.3372864}, it only garnered support from 2 participants in this study.  

\begin{figure}[h]
\centering
\includegraphics[width=0.6\linewidth]{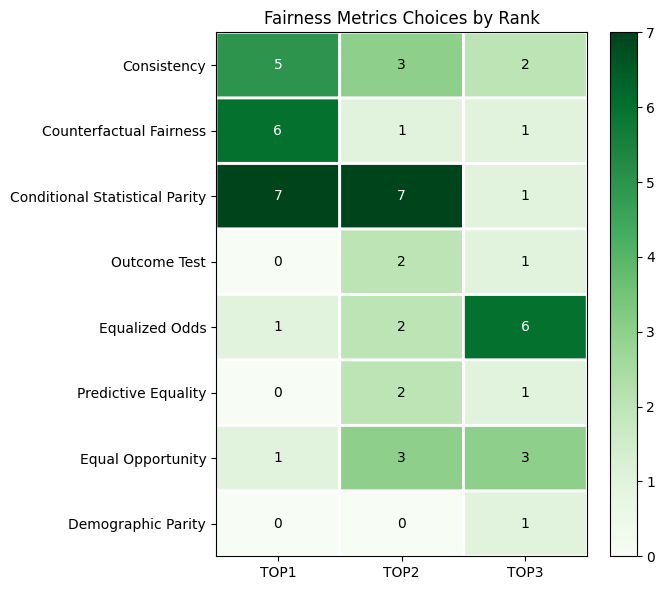}
\caption{The ranking distribution of fairness metrics chosen by participants. The y-axis represents metrics, the x-axis represents three ranks (TOP-1, TOP-2, TOP-3), with color intensity indicating selection frequency. It shows that Conditional Statistical
Parity, Consistency, and Counterfactual Fairness are the most highly ranked metrics.}
\label{fig:metricfrequency}
\end{figure}

Then we delved into specific fairness metrics. Figure~\ref{fig:metricfrequency} shows the ranking distribution of metrics chosen by participants. We observed that preferences diverged among the participants' Top-1 choices. Conditional Statistical Parity was the most preferred, chosen by 7 out of 18 participants, which indicates the importance stakeholders \revision{might} place on non-protected features (job, savings, employment, and credit history in our credit rating scenario) in fairness assessment. It was followed by Counterfactual Fairness (6 participants), and Consistency (5 participants), indicating preferences for fairness at the individual level. It is noteworthy that Demographic Parity, the most widely used metric in existing AI fairness research \cite{hort2023bias,10.1145/3194770.3194776,10.1145/3457607}, was not chosen by any participant as their Top-1 choice. Equal Opportunity and Equalized Odds, which are commonly used fairness metrics, were each selected by only one participant.

\revision{We also observed that there is a wide variability of thresholds in \autoref{tab:FairnessPreference}, driven by a range of \textbf{reasons} (Appendix~\ref{app:Reasons for Fairness Threshold Settings}). No two participants set exactly the same thresholds for all three fairness categories. In group and subgroup fairness settings, only two participants chose a 0\% threshold, while four opted for a 100\% threshold in individual fairness. We can note that on average participants set thresholds relatively stricter than expected by law and by most practices \cite{Gupta2021TransitioningFR}, with a group fairness mean of 9.28\% (SD=7.51\%), a subgroup fairness of 13.00\% (SD=9.96\%), and an individual fairness of 92.44\% (SD=8.53\%).}
 
Most participants were willing to leave \textbf{some room for unfairness}, as it was deemed impossible to achieve in practice. As P14 remarked, "\textit{We don’t need to be too absolute, ... Experts or AI are not 100\% accurate, that absolute perfection in fairness and accuracy is unattainable}". \revision{However, a small number of participants pursued \textbf{perfect fairness}, reflecting a zero-tolerance attitude towards bias.} \revision{Another interesting observation is that participants tended to set more lenient thresholds for subgroup fairness, as \textbf{more features being considered} than in group fairness,} thereby allowing for more leeway. As P10 stated, "\textit{Since you're taking more factors into account when assessing fairness, like more individual factors, you can maybe increase this a bit more, like [...] more leeway}". 

Overall, while there were subgroups of participants that seemed very similar in their preferences, there was a lack of consensus. For example, P10, P11, P16 and separately P6, P14 had identical selections of preferred metrics in all the top three ranks but did not agree with each other. Participants did not agree on fairness metric categories, let alone fairness metrics within them, making \revision{reaching consensus on fairness problematic}.

\vspace{1em}
\noindent
\colorbox{bgcolor}{
  \parbox{\dimexpr\linewidth-2\fboxsep\relax}{
    \textbf{Main finding 1}: \revision{Lay stakeholders' preferences for fairness metrics appeared to differ from the current popular metrics in industry and academia, and more importantly there was no consensus for one metric or metric category.} 
  }
}

\begin{table}
  \begin{minipage}{\textwidth}
  \caption{Participants' Preferences for AI Fairness Metrics and Fairness Thresholds. Top-1, Top-2, and Top-3 represent participants' top three preferred metrics. Group Fairness vs. Subgroup Fairness indicates whether participants prefer applying 
  the selected group-level fairness metric(s) to group fairness or subgroup fairness. Group and subgroup fairness thresholds deem the model fair if differences are below these values. Individual fairness thresholds deem the model fair if results exceed these values.}
  \label{tab:FairnessPreference}
  \resizebox{\columnwidth}{!}{%
  \begin{tabular}{l|llll|lll}
    \toprule
    \multicolumn{5}{c|}{Top 3 Fairness Metric Ranking Lists} & \multicolumn{3}{c}{Fairness Threshold Settings} \\
    \toprule
    Participant ID&Top-1&Top-2&Top-3&\parbox{30mm}{Group Fairness vs. \\ Subgroup Fairness }&\parbox{20mm}{Group Fairness \\ Thresholds} &\parbox{25mm}{Subgroup Fairness \\ Thresholds}&\parbox{25mm}{Individual Fairness \\ Thresholds} \\
    \midrule
    P1 & Conditional Statistical Parity& Equal Opportunity& Counterfactual Fairness& Subgroup& 30\%& 30\%& 70\%\\ 
    P2 & Equailzed Odds& Conditional Statistical Parity& Outcome Test& Subgroup& 20\%& 10\%& 80\%\\
    P3 & Conditional Statistical Parity& Consistency& Predictive Equality& Subgroup& 10\%& 10\%& 90\%\\ 
    P4 & Conditional Statistical Parity& Consistency& Equal Opportunity& Subgroup& 5\% & 5\% & 100\%\\ 
    P5 & Conditional Statistical Parity& Counterfactual Fairness& Demographic Parity& Subgroup& 8\%& 25\%& 100\%\\ 
    P6 & Consistency& Conditional Statistical Parity& Equalized Odds& Subgroup& 5\%& 5\%& 95\%\\ 
    P7 & Counterfactual Fairness& Consistency& Conditional Statistical Parity& Subgroup& 10\%& 10\%& 100\%\\ 
    P8 & Equal Opportunity& Conditional Statistical Parity& Equalized Odds& Subgroup& 2\%& 2\%& 98\%\\
    P9 & Conditional Statistical Parity& Equalized Odds& Consistency& Group& 15\%& 20\%& 95\%\\ 
    P10 & Counterfactual Fairness& Conditional Statistical Parity& Equalized Odds& Subgroup& 10\%& 10\%& 85\%\\ 
    P11 & Counterfactual Fairness& Conditional Statistical Parity& Equalized Odds& Subgroup& 10\%& 20\%& 95\%\\ 
    P12 & Consistency& Outcome Test& Equal Opportunity& Group& 5\%& 20\%& 95\%\\ 
    P13 & Counterfactual Fairness \& Consistency \textsuperscript{1} & \multicolumn{2}{c}{Equal Opportunity/Predictive Equality/Equalized Odds} \textsuperscript{2} & Subgroup& 0\%& 0\%& 100\%\\ 
    P14 & Consistency& Conditional Statistical Parity& Equalized Odds& Subgroup& 10\%& 20\%& 95\%\\ 
    P15 & Conditional Statistical Parity& Predictive Equality& Consistency& Subgroup& 0\%& 0\%& 90\%\\
    P16 & Counterfactual Fairness& Conditional Statistical Parity& Equalized Odds& Depend on Context& 5\%& 5\%& 99\%\\ 
    P17 & Counterfactual Fairness \& Consistency \textsuperscript{1} & Equal Opportunity& Demographic Parity& Group& 5\%& 10\%& 97\%\\ 
    P18 & Conditional Statistical Parity& Outcome Test& Equal Opportunity& Group& 17\%& 32\%& 80\%\\  
  \bottomrule
\end{tabular}
}
\footnotesize
\noindent\textsuperscript{1} P13 and P17 saw Counterfactual Fairness and Consistency as equally important, and could not distinguish between them in terms of ranking. \\
\textsuperscript{2} P13 preferred to choose the metric based on the proportion of rated credit labels: Equal Opportunity when "Good" is the majority, Predictive Equality when "Bad" is the majority, and Equalized Odds when both are equal. \\
\normalsize
\end{minipage}
\vspace{-1em}
\end{table}

\subsubsection{Reasons for Metric Preferences and Concerns}
\label{sec: reasons for fairness preferences}
We analyzed participants' think-aloud data from individual sessions using a codebook (Appendix~\ref{app: Participants' Reasons for Selection and Exclusion}) and highlighted the code in this section in bold text. We expected these reasons to be reprised and discussed during group negotiations as arguments for a metric consensus, \revision{as detailed in Section~\hyperref[sec:negotiation]{5.2}}.

Recall that individual fairness was the most favored fairness category. Nine participants who opted for individual fairness did so because they placed importance on \textbf{individual differences} which group-level fairness, in their opinion, overlooks. As P10 stated, "\textit{It's like you're looking at the individual themselves, […] instead of the group features, which may differ based on, you know, individual differences}". 

If group fairness metrics were selected, participants also specified whether subgroup fairness was more important. Only 4 participants opted for group fairness, emphasizing a \textbf{focus on one protected feature}. They argued that fairness, such as gender equality, should be evaluated without the interference of other factors: \textit{"If we care about gender equality, we only focus on female and male, gender. This is more focused, we should not let other factors influence our judgment if we achieve gender equality or not" (P9)}. The majority (13 participants) explicitly preferred subgroup fairness. They suggested that subgroup fairness was an option that to a certain extent combined individual and group fairness, minimizing group differences by considering multiple protected features in a \textbf{macro-to-micro perspective}: "\textit{With subgroups it's almost like looking from a macro to a microscale. When you're looking on a group basis, the whole point of a subgroup is because there's some similarities within the group,[…] so when you're looking at subgroup basis, you can minimize the differences more. If you minimize differences in in the group. You may still have differences in the subgroup}" (P10).

We then looked into the reasons and concerns for the most highly ranked metrics, which were Conditional Statistical Parity, Consistency, and Counterfactual Fairness. For \textbf{Conditional Statistical Parity}, 7 participants highlighted \textbf{important non-protected features} for them, such as employment, savings, and job, recognizing their value for both accuracy and fairness and preferring using these when dividing groups. As P10 said, "\textit{They're all predicting factors}". Five participants believed that \textbf{non-protected feature-specific comparisons} are fairer, as they reduce the risk of introducing other forms of discrimination. As P6 put it, "\textit{I have a management job, I will care more about the people who also have a management job. […] This is fairer. Otherwise, people with different job types will be mixed together and then calculate fairness. There's another discrimination […]. You should not let other factors influence too much}". Similarly, participants emphasized that grouping people based on important non-protected features is \textbf{objective}, making fairness assessment \textit{"about the ability of each person, regardless of their characteristics" (P18)}. Additionally, three participants thought that the metric allows for \textbf{flexible bias detection} by enabling further subdivision of people into groups based on different conditions for a more targeted analysis: "\textit{It's that it can include various factors. It can be more targeted and flexible (P5)}". However, one participant (P17) noted the \textbf{non-protected features' susceptibility} in addressing biases because they are often related to protected features: "\textit{There's a significant difference between two age groups, like in credit history and employment. Age could influence this, relationships. So, dividing them under the same condition isn't fair}".

For \textbf{Consistency}, five participants argued that it allowed \textbf{comparison with similar individuals}: "\textit{If people similar to me predictably have good credit, I should also receive a favorable prediction. It's fair. It allows you to compare yourself with people of the same level or qualifications}" (P14). Additionally, P12 mentioned that this comparison provides a degree of \textbf{decision transparency}, helping to understand the prediction result: "\textit{I can compare to others to understand why I cannot achieve better results"}. Meanwhile, one participant believed that it provides \textbf{inclusivity for individuals}, ensuring no one is unfairly excluded, for example, \textit{"'Cause it prevents discrimination like, no one's getting left out sort of thing [among similar individuals] (P13)}". Nonetheless, five participants expressed concerns over a downside: how \textbf{similarity is quantified} could be biased, complicating the definition of similar individuals. As P16 noted, "\textit{There is no way or hard to find a standard way to analyze and compare similarity objectively}". 

In terms of \textbf{Counterfactual Fairness}, five participants chose this metric because they believed it ensures \textbf{irrelevance of protected features}, such as age, gender, or foreign worker status, as exemplified by P17, "\textit{Because it disregards their age, their gender and their foreign worker status. […] Nothing should affect it at all}". Moreover, four participants favored its \textbf{individual characteristic-based consideration}, specifically analyzing if protected features affect one specific individual rather than conducting broader, general group-based assessments, for instance, \textit{"The factors that are that mostly contribute to a bias […] you're looking at it on an individual basis. […] You take their individual characteristics into account (P10)"}. Last, compared to Consistency, three participants found the \textbf{objective quantification} of Counterfactual Fairness convincing because \textit{"it is like I can quantity the similarity because the overall information for this person is still not changed, we just change the personal characteristics (P5)}". Nevertheless, P2 expressed reservations regarding its \textbf{practical ineffectiveness} compared with group fairness, highlighting the minimal effect of modifying a feature value, especially given that protected feature weights are nearly zero: \textit{"Because the digital I think they are mostly the same, and however, I don't think there are so many big, big differences"} after altering protected feature values, yet adding, \textit{" [but] I can see the difference between the different age people [group]"} (P2).

\vspace{1em}
\noindent
\colorbox{bgcolor}{
  \parbox{\dimexpr\linewidth-2\fboxsep\relax}{
    \textbf{Main finding 2}: \revision{There were numerous justifications for choosing particular metrics among participants.  Participants who preferred individual fairness often pointed to individual differences overlooked by group-level metrics. Participants also placed importance on non-protected features that influence AI fairness decisions and emphasized evaluations based on capabilities, which motivated them to choose Conditional Statistical Parity. Participants who chose Consistency noted that it allowed comparisons among similar people, promoting fairness and inclusivity. Counterfactual Fairness was praised for concentrating on individual traits and objective quantification}.
  }
}

\subsection{How do stakeholders negotiate to resolve personal differences in metric preferences? (RQ2)}
\label{sec:negotiation}
Because stakeholders have diverse views and preferences on metrics and fairness categories, reflecting these views in "fairer" AI models remains challenging. In this section, \revision{we investigate how team members negotiated to resolve differences in metric preferences after using FEE to review and build a shared understanding of each selected metric. We identified three distinct negotiation strategies for consensus-reaching, achieved without researcher input.} We will detail the negotiation strategies shaped by team decision-making and leadership, and the consensus each team ultimately reached.

In \textbf{Team 1}, participants adopted mainly a majority-based voting approach to resolve their personal metric preferences. In this approach, they narrowed down options based on the frequency of preferred metrics. Team members agreed to follow the majority and abandon any minority preferences. They started out by choosing to prioritize subgroup fairness over group fairness, following a majority vote suggested by P11. Next, they eliminated the three least chosen metrics, as suggested by P8. Further, as proposed by P4, they narrowed down the selection of the specific metric to employ to the Top-1 choice, either Conditional Statistical Parity or Counterfactual Fairness. At this point, P7, P11, and P12 favored Counterfactual Fairness due to their personal Top-1 preference for individual-level fairness, while P4, P5, and P8 supported Conditional Statistical Parity, driven by their Top-1 preference for group-level fairness. This then led to in-depth discussions between team members as to the reasons behind their Top-1 choices. \revision{During the discussion, team members consulted FEE for insights on Conditional Statistical Parity and Counterfactual Fairness, resurfacing and sharing what they had talked about during their earlier individual sessions (Section~\hyperref[sec: reasons for fairness preferences]{5.1.2}).} We also found that the process of discussion at this stage stimulated the exploration of the core advantages and disadvantages of fairness and ways to measure it, thus deepening participants' understanding of the metrics. For example, P7 supported Counterfactual Fairness and pointed out the drawbacks of group fairness: "\textit{The thing is the subgroups just level up other groups, right? So if you and I have the same credit score and we apply for a credit, […], as I said, I don't get the credit but you get it because the metrics is 50\% of all males good credit, only 30\% of the females good credits. So we have to give more credits to females. So we have exactly the same income, but [you]'ll be prioritized because you belong to a group [...] [What] group and subgroup [metrics] generally do is to try from a government perspective, equalized. But the question is, is it fair or not?}". On the other hand, P8, a supporter of Conditional Statistical Parity, expressed concerns about individual fairness, stressing the importance of group fairness in maintaining equity between protected and non-protected groups at the societal level: "\textit{It's maybe unfair [...] [Unless] I'm selfish, I will go with individual fairness}". Meanwhile, P11 emphasized the decision not to use Consistency due to the complexity of similarity calculations, explaining, '\textit{Instead of neighbor because you have to explain again what this neighbor means, it is that you have the same [information]. But if the counterfactual fairness, it's just like male [or] female [...] or something like that}'. This nuance was not fully realized during P11's individual session, yet group discussions further deepened the understanding of this metric. Therefore, after this discussion, the team agreed on a "hybrid" solution: \textbf{using Conditional Statistical Parity with a focus on subgroup fairness first, and then using Counterfactual Fairness as a backup for individual fairness}. As P8 suggested: "\textit{[Use] Conditional Statistical Parity first and then I when I see a complaint then I would go with the Counterfactual Fairness to decide to, to judge him a second time}".

\textbf{Team 2}, similar to Team 1, also tried to discuss the reasons for their choices and then vote on them, emphasizing collective agreement to reconcile their diverging preferences. However, this discussion led to entrenched positions and inhibited negotiation. For example, initially, P1 suggested that each team member share and justify their top 3 preferred metrics in turn. \revision{In this process, each one used FEE to demonstrate their metric preferences.} Then, they checked if any team member had changed their mind after hearing the others. However, all members stuck to their initial personal choices. Then, as suggested by P17, they tried to overcome that deadlock by voting on the team's top 3 metric list. However, the team only accepted unanimous decisions and therefore this approach failed. As a final way out, the team attempted a compromise solution. P17 suggested: "\textit{I like the idea of grouping with the Conditional Statistical Parity and then choosing an individual fairness afterward. And within same condition groups that we want to make sure each individual can be treated fairly}". All members, except P15, favored individual fairness with Counterfactual Fairness, while P15 suggested a combination of Counterfactual Fairness and Consistency. For comprehensiveness, they chose to combine both individual fairness metrics. Team 2 further prioritized subgroup fairness for Conditional Statistical Parity using a majority-based approach. Therefore, similar to Team 1, Team 2 also agreed on a "hybrid" solution: \textbf{using Conditional Statistical Parity with a focus on subgroup fairness first, and then using Counterfactual Fairness and Consistency as backups for individual fairness}. 

In \textbf{Team 3}, reasons for and against fairness metrics were also discussed but they used a debate style, encouraging the presentation of arguments for and against certain fairness metrics. In the early negotiation stages, some polarization emerged similar to Team 1, but was successfully overcome through argumentation. For example, P10, P13, and P14 supported Counterfactual Fairness, while P9 and P18 favored Conditional Statistical Parity. In this team, we observed that certain team members dominated the discussions, leading to significant variations in the levels of participation among the group members. P10 presented arguments for Counterfactual Fairness while P18 supported Conditional Statistical Parity, leading two sides. The discussion primarily revolved around P18's focus on group-level legitimate features for fairness versus P10's approach of directly excluding the impact of protected features for AI decision-making fairness, \revision{consulting FEE as needed to support their arguments}. P10 proposed fairness as follows: "\textit{It's like oh no matter the value of age, gender, foreign worker is changed or not, AI prediction should not be changed […] You eliminate the bias first and then you think about employment [and other individual qualifications]}". After this perspective was accepted, Team 3, different from Team 1 and Team 2, advocated for \textbf{using Counterfactual Fairness for individual fairness first, and then using Conditional Statistical Parity with a focus on subgroup fairness}. Additionally, there was unanimous agreement on the importance of including at least age and gender to secure subgroup fairness in the Conditional Statistical Parity application.  

\vspace{1em}
\noindent
\colorbox{bgcolor}{
  \parbox{\dimexpr\linewidth-2\fboxsep\relax}{
    \textbf{Main finding 3}: Teams used different strategies to come to metric consensus. Voting was frequently employed but there were differences in how to decide the outcomes of voting. Reasons for or against metrics were discussed in all teams but this could also lead to entrenchments of opinions if not carefully handled. All teams chose to overcome conflict by combining metrics in some way, ending up with compromise solutions. The consensus showed that all teams considered both subgroup fairness and individual fairness, but with different priorities.
  }
}

\section{Discussion}
\label{sec: Discussion}
In this section, we will first discuss the limitations of our work and future research. \revision{Then, we will discuss the implications for the design of fairness tools and processes by considering how UIs could be extended to express personal preferences, the complexity of accommodating diverse personal fairness preferences, how to support negotiation in groups, and how to alleviate problems in reaching consensus. We reflect on how these tools and processes would need to scale up to take into account larger and distributed stakeholder groups.}

\subsection{Limitations and Future Work}
We acknowledge four limitations of our work. First, the current sample of participants in the user study was relatively small and focused solely on lay users as decision subjects without involving other stakeholder types, such as policymakers or domain experts. Future work needs to investigate how our framework could be scaled up and include multiple stakeholder types. Second, we only implemented a subset of fairness metrics predefined by AI experts. Future work could expand the range of metrics or induce custom fairness metrics from users to better reflect their understanding and expectations of fairness. Third, although we attempted to provide a comprehensive UI based on best practices, there might have been some limitations to its usability. Participants' feedback on UI design was coded into themes (in bold), see Appendix \ref{app:Participants' Feedback on UI Design}. Although FEE received many positive comments from participants, including features such as \textbf{real-time updates to fairness results} (8 participants) and \textbf{intuitive visualization for fairness metric explanations} (6 participants) to bridge lay users' AI knowledge gap, it was \textbf{cognitively challenging} (8 participants), as P5 stated: \textit{"Mind-twisting when I compare them"}. Future work could concentrate on explaining the \textbf{practical implications of fairness metrics} (6 participants) by clearly linking calculations to real-world scenarios. Additionally, adopting \textbf{simplified data visualization \& explanation} (3 participants) to explain metric principles and \textbf{model transparency} (2 participants) is crucial for understanding decision-making reasons. Last, we only tested our framework and UI in the credit rating scenario, limiting the scope of our findings. Future work should evaluate and refine our framework in diverse settings like healthcare, employment, and education to ensure its broader applicability and effectiveness in addressing fairness. 

\subsection{\revision{Implications for Human-Centered Fairness}}
We argue that tackling AI fairness transcends the bounds of solely technical, AI expert-driven approaches, evolving into a broader, human-centered concern. Our work demonstrates the feasibility of \revision{this} approach. We will discuss \revision{how our work presenting the EARN Fairness framework and the FEE interactive system has implications for design to support stakeholders' elicitation of personal preferences and negotiation toward fairness metric consensus}.

\subsubsection{Explaining Fairness Metrics}
As mentioned in Section~\hyperref[sec: Related Work]{2}, most research indirectly explores metrics that resonate with users' fairness perceptions \cite{10.1145/3292500.3330664,10.1145/3514258, umapworkshop,10.1145/3306618.3314248}. Instead, our framework shares a similar philosophy with the work of Cheng et al. \cite{10.1145/3411764.3445308}, using the FEE interactive system to reveal the underlying logic of metrics, and thus enabling users to decide which fairness metric aligns best with their expectations. Although our FEE tool only covered a subsection of metrics, this could be easily extended to cover more and different fairness metrics, for a range of features extending beyond protected and legitimate features. Our tool also allows researchers to extend it to different application scenarios by changing datasets or incorporating other AI models' fairness outcomes.

Explaining fairness metrics through FEE also had the effect of engaging the participants in our study to think about fairness, how fairness is defined, and what fairness means to them. For example, participants excluded metrics that define fairness within specific groups, such as Equal Opportunity, which ensures fairness only within the subgroup labeled "Good Credit". Furthermore, we observed that when participants made modifications to AI predictions or ground-truth labels to explore real-time updates in fairness results, it prompted them "\textit{reflect or re-think how the fairness results changed}" (P14), enhancing their understanding of the metrics. Therefore, we believe that soliciting stakeholders' metric preferences based on exposing the underlying logic offers a robust approach. 

\subsubsection{\revision{Complexity of Diverse Fairness Preferences}}
From the perspective of personal preferences, \revision{we observed differences between the fairness metrics preferred by stakeholders without AI expertise and those commonly used by AI experts in industry and academia}, such as Demographic Parity, Equal Opportunity, and Equalized Odds \cite{xu2022ai,financefairness_recent_2023}. Similarly, Nakao et al. \cite{doi:10.1080/10447318.2022.2067936} found that loan officers' preferences diverged from data scientists'. This highlights the need to involve a broader base of stakeholders in fairness metric decisions. However, even within the same type of stakeholders, in this case, decision subjects, we observed different understandings of fairness, not only in metrics but also in fairness categories. This finding echoes previous research \cite{10.1145/3334480.3375158,landers_auditing_2023,10.1145/3411764.3445308}. 

This leads to the question of what to do with \revision{different} personal preferences once we have elicited them, in order to \revision{accommodate differences. Given that there is no "one-size-fits-all" metric and selecting only one metric is insufficient \cite{10.1145/3457607}, one possible approach for AI practitioners and researchers, as revealed through our findings, is \emph{preference integration}.} When teams negotiated their preferred fairness metric, we observed a common pattern: all teams tended to merge diverse preferences to form a more comprehensive and inclusive definition of AI fairness to resolve differences, despite using different negotiation strategies. \revision{That is, in the credit rating scenario, all teams agreed on a dual focus on the absence of prejudice or favoritism toward both individuals and groups}. While individual and group fairness might not inherently conflict as they reflect different aspects of consistent moral and political concerns \cite{10.1145/3351095.3372864}, conflicts may arise in practice when optimizing specific metrics \cite{10.1145/3457607}. \revision{We will next discuss how negotiations around conflicting personal preferences could be supported.} 

\subsubsection{Supporting Team Negotiation} 
In our framework, we emphasized collective metric consensus to foster mutual acceptance, and our findings demonstrate that stakeholders can reach a consensus without the interference and aid of researchers. \revision{We observed that varying strategies were adopted for negotiating consensus within the teams of our study, indicating that robust negotiating approaches need to be developed, codified, and supported by tools and processes.}

\revision{Our current interactive system, FEE, allows each user to access fairness metrics, track assessment results, and further helps them rank metrics and determine fairness thresholds. While it can be used during the group negotiation phase to review metrics together, it does not support any structured group negotiation.}

\revision{We propose a series of design improvements to support better group negotiation and communication. First, tools such as FEE will need to be expanded to display personal fairness metric preferences of individuals, as shown by other tools used in algorithmic fairness \cite{10.1145/3359284}. This will enable stakeholders to see others' choices, helping them better understand other positions. It remains to be investigated what the implications are of making these choices directly attributable to individuals or only making anonymous contributions visible.}

\revision{Differences in choices could be made visible through visualization tools to facilitate collective decision-making. For example, tools such as graph visualization can display the proximity of individual preferences \cite{graphvisualization}, while visual ratings of support or opposition to specific metrics \cite{10.1145/2675133.2675272,10.1145/2145204.2145249} can quickly pinpoint areas of potential consensus or disagreement. Meanwhile, it needs to be ensured that while guarding against discriminatory views, minority opinions are heard and the negotiation process is fair.}

\revision{Therefore, users should be able to provide a brief justification for their selected fairness metrics. This would foster an evidence-based negotiation environment where stakeholders can engage in meaningful discussions. The negotiation process itself could then be enhanced by incorporating tools such as reason-based argumentation frameworks \cite{reasonbasedargumentation}. This approach formally models moral debates, meticulously evaluates the moral acceptability of each argument, and clearly demonstrates the logical foundation and interrelations among them, thus ensuring the negotiation process is fair, inclusive, and conducive to building consensus.}

\subsubsection{\revision{Handling Deadlocks and Failures During Negotiations}}

\revision{We also observed instances of temporary "voting deadlock" during group negotiations but there might also be more permanent failures to reach consensus employing the EARN Fairness framework.} 

\revision{First, rules for building consensus could be embedded into tools that support negotiation, such as voting mechanisms \cite{10.1145/3359284, 10.1145/3359283},  including majority-based voting observed in our group sessions, or the Borda voting method, as implemented in the "WeBuildAI" framework \cite{10.1145/3359283}. Additionally, tools could support game-rule-driven negotiations to reduce the risk of failure \cite{10.1145/3544548.3581527}. Gamified elements can enhance team cohesion, promote group identity, stimulate group dynamics, and improve enjoyment and user experience \cite{10.1145/3359159}. However, when designing the "game rules" for negotiation, it would be essential to encourage cooperative rather than competitive engagement, to avoid personal preferences becoming entrenched, akin to what we saw in Team 2 (Section~\hyperref[sec:negotiation]{5.2}).}

\revision{To overcome permanent failures, we might also consider optimizing \emph{across all} metrics chosen by stakeholders. To do this we can apply a preference weighting mechanism to calculate the score for each metric based on individual preferences, which automatically recommends the "winning" metric. For example, in our case, weights of 3, 2, and 1 can be assigned for the Top-1, Top-2, and Top-3 choices, and a final weighted ranking can be displayed to the users. In our credit rating scenario, Conditional Statistical Parity (36 points) far outweighed Consistency (23 points), Counterfactual Fairness (21 points), and other metrics. However, in our view, these approaches relate back to attempting to solve a socio-technical problem with more technical solutions.} 

\subsubsection{Scaling Up the Process} Our work only involved 18 individuals, in three small teams of 6 stakeholders. This raises the question of how to scale up our framework to work with larger numbers of stakeholders in their 100s or even 1000s.  

The first part of the EARN \revision{Fairness} process, the individual sessions, could be readily translated online, and used by people without researcher involvement in an asynchronous and distributed manner. This relies on the stand-alone use of the FEE tool, and choosing features and at least one fairness metric. 

\revision{To support negotiation among distributed groups of people, developing an online system for a larger group is essential. This could be achieved by integrating synchronous and asynchronous online collaboration tools. For example, other tools \cite{10.1145/3579601, 10.1145/3613904.3642849} allow personal preference data to be imported into online whiteboards to display the results to a group of stakeholders and attach reasoning or justifications, thus providing a foundation for deliberation. However, these tools may not be adequate for larger groups, as negotiation support can quickly become unwieldy with too many participants. How to address this challenge is an open question, but it could be solved with other democratic processes already in place, for example, by choosing stakeholder representatives.}

\section{Conclusions}
In this paper, we proposed the EARN Fairness framework to uncover fairness metric consensus among stakeholders without prior AI knowledge. We applied our framework in a credit rating scenario with 18 participants to identify decision subjects' personal metric preferences\revision{, and then, by grouping them into three teams of 6 participants each, we studied how they negotiated consensus on metric selection}. We found that:

\begin{itemize}
    \item \revision{Lay stakeholders' metric preferences differ from popular metrics in industry and academia, and there is no overall preference for one metric. There is a need for consensus to be negotiated among stakeholders.}
    \item \revision{There are many, perfectly good reasons for choosing fairness metrics which are used in negotiating fairness in groups.}
    \item \revision{We observed some problems in reaching consensus during the negotiations, mainly centering on ways to resolve conflicts. For consensus, groups typically choose to compromise by combining multiple metrics to address differences in personal preferences.}  
\end{itemize} 

\revision{We discussed the implications for future tools and processes that support reaching consensus on choosing fairness metrics. Our framework, comprising the EARN Fairness process and the interactive system, presents a feasible step for engaging stakeholders in the expression of personal preferences and collective selection of metrics, fostering more equitable and inclusive AI fairness}.

\begin{acks}
   We gratefully acknowledge the funding provided by the University of Glasgow and Fujitsu Limited. We thank all participants for their contributions.
\end{acks}

\bibliographystyle{ACM-Reference-Format}
\bibliography{EarnFairness}

\clearpage
\appendix
\section{Appendix}

\subsection{Fairness Metrics and Explanations}
\label{app:Fairness Metrics and Explanations}
\begin{table}[H]
  \caption{Fairness Metrics and Explanations}
  \label{tab:Metrics}
  \captionsetup{skip=0pt}  
  \small
\resizebox{\columnwidth}{!}{%
  \begin{tabular}{|p{0.07\columnwidth}|p{0.1\columnwidth}|p{0.04\columnwidth}|p{0.27\columnwidth}|p{0.52\columnwidth}|}
    \hline
    Category&Metric&Paper&Description &Calculation \\
    \hline
    \multirow{6}{=}{\parbox{0.07\columnwidth}{\textsuperscript{1}Group Fairness}} & Demo-graphic Parity & \cite{NIPS2017_a486cd07}, \cite{10.1145/2090236.2090255} & Both protected and unprotected group members have an equal likelihood of being classified in the positive category. & \text{Difference} = $P(\hat{Y} = \text{1} \mid \text{G = 0}) - P(\hat{Y} = \text{1} \mid \text{G = 1})$ \\
    \cline{2-5}
    & Equal Opportunity & \cite{10.5555/3157382.3157469} & Both protected and unprotected groups have the same probability that a subject in the positive class is assigned a positive predictive value. & \text{Difference} = $P(\hat{Y} = \text{1} \mid \text{Y = 1, G = 0}) - P(\hat{Y} = \text{1} \mid \text{Y = 1, G = 1})$ \\
    \cline{2-5}
    & Predictive Equality & \cite{10.1145/3097983.3098095} & Both protected and unprotected groups have the same probability that a subject in the negative class is assigned a positive predictive value & \text{Difference} = $P(\hat{Y} = \text{1} \mid \text{Y = 0, G = 0}) - P(\hat{Y} = \text{1} \mid \text{Y = 0, G = 1})$  \\
    \cline{2-5}
    & Outcome Test & \cite{articleoutcometest} & Both protected and unprotected groups have the same probability that a subject with a positive predictive value genuinely belongs to the positive class. & \text{Difference} = $P(\text{Y = 1} \mid \hat{Y} \text{= 1, G = 0}) - P(\text{Y = 1} \mid \hat{Y} \text{= 1, G = 1})$  \\
    \cline{2-5}
    & Equalized Odds & \cite{10.5555/3157382.3157469} & Simultaneously satisfy both Equal Opportunity and Predictive Equality. & 
    
    \text{Difference1} = $P(\hat{Y} = \text{1} \mid \text{Y = 1, G = 0}) - P(\hat{Y} = \text{1} \mid \text{Y = 1, G = 1})$
    
    \text{Difference2} = $P(\hat{Y} = \text{1} \mid \text{Y = 0, G = 0}) - P(\hat{Y} = \text{1} \mid \text{Y = 0, G = 1})$
    
    \text{Difference} = $\max(\text{Difference1}, \text{Difference2})$ \\
    
    \cline{2-5}
    & Conditional Statistical Parity & \cite{10.1145/3097983.3098095} & Subjects in both protected and unprotected groups are equally likely to be classified into the positive predicted class, taking into account a set of legitimate features L. In our context, to facilitate lay users' understanding, we adopted four non-protected features (i.e., job, savings, employment, and credit history) and calculated the metric value separately for each. & \text{Difference} =$P(\hat{Y} = \text{1} \mid \text{L = l, G = 0}) - P(\hat{Y} = \text{1} \mid \text{L = l, G = 1})$  \\
    \hline
    \multirow{2}{=}{\parbox{0.07\columnwidth}{Indivi-dual Fairness}} & Counter-factual Fairness & \cite{NIPS2017_a486cd07} & An individual receives the same decision in both the actual world and a counterfactual world where the individual belonged to a different demographic group. & 
    \(
\textsuperscript{2}CFR = \frac{1}{N} \sum_{(X, A_{\text{true}}) \in \text{dataset}} 1 \{p(X, A_{\text{true}}) = p(X, A_{\text{counterfactual}})\}
\) 
 \\
 \cline{2-5}
    & Consistency & \cite{pmlr-v28-zemel13} & Similar individuals receive similar outcomes. & \(
\textsuperscript{3}\text{Consistency} = 1 - \frac{1}{n} \sum_{i=1}^{n} | {\hat{y_i}} - \frac{1}{n\_\text{neighbors}} \sum_{j \in \mathcal{N}_{n\_\text{neighbors}}(x_i)}^{} \hat{y_j}| \) \\
 \hline
\end{tabular}
}
\end{table}

\footnotesize
\noindent\textsuperscript{1} The group fairness Metric results are represented by the difference between groups. \\
- $\hat{Y}$: AI-predicted credit rating. $\hat{Y} = 1$ denotes a prediction of good credit, while $\hat{Y} = 0$ denotes a prediction of bad credit. \\
- $Y$: Actual credit rating label. $Y = 1$ indicates a good credit rating as per the original data, while $Y = 0$ indicates a bad credit rating.\\
- $G$: Group membership variable. $G = 0$ denotes the protected group, while $G = 1$ denotes the non-protected group. \\
Similarly, the subgroup fairness metric results are represented by the maximum difference between any two subgroups. \\
\noindent\textsuperscript{2} Counterfactual fairness is quantified by the proportion of instances where decisions remain consistent between the actual and counterfactual worlds. \\
- \(N\) represents the total number of instances in the dataset. \\
- \(X,A_{\text{true}}\) denotes an original instance \(X\) with the actual protected feature value \(A_{\text{true}}\) \\
- \(X,A_{\text{counterfactual}}\) denotes an original instance \(X\) with the counterfactual protected feature value \(A_{\text{counterfactual}}\) \\
- \(p(X, A_{\text{true}})\) presents the model's prediction for the instance \(X,A_{\text{true}}\) \\
- \(p(X, A_{\text{counterfactual}})\) presents the model's prediction for the instance \(X,A_{\text{counterfactual}}\) \\
- \(1\{condition\}\) is an indicator function, returning 1 if the given condition is true, otherwise returning 0. \\
\noindent\textsuperscript{3} The calculation of Consistency score is computed based on the similarity of AI predictions among the five nearest neighbors, as per IBM's AI Fairness 360 \cite{8843908}. \\
\normalsize

\subsection{Participant Information}
\label{app:Participant Information}
\begin{table}[H]
  \caption{Participant Information}
  \label{tab:Participant}
  \vspace{-3mm}
  \scalebox{0.8}{
  \begin{tabular}{cclllll}
    \toprule
    Participant ID&Age&Gender &Education &Profession &Nationality &Ethnicity\\
    \midrule
    P1& 47& Male& Master& Facilities& British&White\\
 P2& 18& Female& High School& Computer and IT& Chinese&Asian\\
    P3& 28& Male& Bachelor& Healthcare and Medical& Chinese&Asian\\
    P4& 23& Female& Bachelor& Business and Financial& Chinese&Asian\\
    P5& 25& Female& Bachelor& Business and Financial& Chinese&Asian\\
 P6& 23& Female& Bachelor& Healthcare and Medical& Chinese&Asian\\
 P7& 31& Male& Doctorate& Education& Austrian&White\\
 P8& 25& Male& Master& Business and Financial& Iranian&Asian\\
 P9& 29& Female& Bachelor& Hospitality and Tourism& Indonesia&Asian\\
 P10& 26& Male& Bachelor& Healthcare and Medical& Bangladeshi&Asian\\
 P11& 43& Female& Master& Business and Financial& Indonesia&Asian\\
 P12& 29& Female& Master& Business and Financial& Chinese&Asian\\
 P13& 23& Male& Master& Computer and IT& British&Arab\\
 P14& 26& Female& Master& Sociology& Chinese&Asian\\
 P15& 25& Male& Bachelor& Engineering& Indian&Asian\\
 P16& 33& Female& Doctorate& Art& Chinese&Asian\\
 P17& 27& Female& Master& Facilities& Jamaican&Black\\
 P18& 21& Male& Bachelor& Biotechnology and Management& Indian&Asian\\
 \bottomrule
\end{tabular}
}
\end{table}
\clearpage 

\subsection{Codebooks}
\label{app:codebook}
\subsubsection{Reasons for Fairness Threshold Settings}
\label{app:Reasons for Fairness Threshold Settings}
We analyze the reasons behind participants' threshold settings.
\begin{table}[H]
  \caption{Codebook-Reasons for Fairness Threshold Settings}
  \label{tab:Codebook Thresholds}
  \resizebox{\columnwidth}{!}{%
    \begin{tabular}{|p{0.25\textwidth}|p{0.3\textwidth}|p{0.35\textwidth}|p{0.05\textwidth}|p{0.05\textwidth}|}
    \hline
    Code&Code Description&Example&Pts. &Freq.\\
    \hline
    Perfect Fairness & Expressing the preference for equal treatment of all individuals or groups, ensuring no disparities exist among them. & According to the definition of individual fairness, each individual should be treated fairly. (P5)
No difference between any groups. (P15)&5&6\\
\hline
 Some Room for Unfairness& Expressing the practical concerns that absolute perfection in fairness and accuracy is unattainable, and a small margin for exceptions or unfairness.& I think there is some space. (P4) The less the better but 0\% or 100\% is impossible. (P6) We don’t need to be too absolute, ... Experts or AI are not 100\% accurate, that absolute perfection in fairness and accuracy is unattainable. (P14)
& 10&16\\
\hline
  Varying Threshold Dependent on Protected Features& Expressing flexibility in fairness thresholds, adapting the tolerance levels to the specific protected features, such as age or gender.& Maybe different percentage in different under different criteria. (P4)
& 1&1\\
\hline
 More Features Being Considered& Expressing a looser fairness threshold when accounting for a broader range of individual factors, facilitating essential trade-offs.& Since you're taking more factors into account when assessing fairness, like more individual factors, you can maybe increase this a bit more, like [...] more leeway. (P10)
& 4&4\\
\hline
 Trade-off between Fairness and Accuracy& Expressing the belief that enhancing AI fairness might impact accuracy, setting thresholds to balance this trade-off.& That is in the sense if you're being more fair then the model is getting more inaccurate,  but if it's less fair then it means it's more accurate which I think we are. … So you have to improve the fairness. But at the same time we have to make sure that the accuracy of the Ai's system. I have to balance them both. (P18)
& 1&5\\
\hline
 Law Requirements& Expressing adherence to legal thresholds and fine-tuning thresholds within legal guidelines to evaluate the trade-offs between potential gains and losses.& We could choose the threshold according to the law. [...] Then I would just tune up, tweak like if we go up (threshold) this much, how much do I gain versus how much do I lose.(P13)
& 1&2\\
 \hline
\end{tabular}
}
\end{table}
\clearpage

\subsubsection{Participants' Reasons for Selection and Exclusion}
\label{app: Participants' Reasons for Selection and Exclusion}
We analyze the data from two perspectives: fairness categories and specific metrics.

\begin{table}[H]
  \caption{Codebook-Fairness Category}
  \label{tab:Codebook Fairness Requirements}
   \captionsetup{skip=0pt} 
  \resizebox{\columnwidth}{!}{%
    \begin{tabular}{|p{0.1\textwidth}|p{0.25\textwidth}|p{0.25\textwidth}|p{0.3\textwidth}|p{0.05\textwidth}|p{0.05\textwidth}|}
    \hline
     Subtheme&Code&Code Description&Example&Pts. &Freq.\\
    \hline
     Prefer Individual Fairness &Individual Differences  & Expressing users' perspective that individual fairness is inherently the fairest, emphasizing users' preference for ensuring fairness at the individual level instead of the group level. & It's like you're looking at the individual themselves, […] instead of the group features, which may differ based on, you know, individual differences. (P10)&9&13\\
    \hline
  Prefer Group Fairness  &Focus on one protected feature & Expressing users’ preference for exclusive focus on a single protected feature for fairness, minimizing the influence of other factors. & If we care about gender equality, we only focus on female and male, gender. This is more focused, we should not let other factors influence our judgment if we achieve gender equality or not. (P9)
& 4&5\\
\hline
   \multirow{3}{=}{\parbox{0.1\textwidth}{Prefer Subgroup Fairness}}&Holistic Fairness Across Multi-Factors & Expressing users’ preference for achieving comprehensive fairness by encompassing all relevant protected features, such as age and gender. & Combine everything together. It shouldn't. Just don't take one part of it into account. It should be to make it fairer. (P1)
   
   Not only one specific features that's also the combined features. (P11)
& 3&3\\
\cline{2-6}
   &Macro-to-Micro Perspective &Expressing that users support micro fairness through refined subgroup analysis, facilitating the identification of more biases and minimizing differences among subgroups. & With subgroups it's almost like looking from a macro to a microscale. When you're looking on a group basis, the whole point of a subgroup is because there's some similarities within the group,[…] so when you're looking at subgroup basis, you can minimize the differences more. If you minimize differences in in the group. You may still have differences in the subgroup. (P10)
& 5&5\\
\cline{2-6}
   &Flexibility in Bias Detection & Expressing that users’ preference group fairness due to allowing more flexible bias detection through customized feature selections. & For subgroup fairness, I can select more than one features and less than the all features. (P3)
& 1&1\\

\hline
\end{tabular}
}
\end{table}
\clearpage

\begin{table}[H]
  \caption{Codebook-Reasons for Selection and Exclusion}
  \label{tab:Codebook-SlectionExclusion}
  \captionsetup{skip=0pt} 
  \resizebox{\columnwidth}{!}{%
    \begin{tabular}
	{|p{0.13\textwidth}|p{0.1\textwidth}|p{0.22\textwidth}|p{0.2\textwidth}|p{0.25\textwidth}|p{0.05\textwidth}|p{0.05\textwidth}|}
  	\hline
  	Theme & Subtheme & Code & Code Description & Example & Pts. & Freq. \\
  	\hline

    \multirow{6}{=}{\parbox{0.1\textwidth}{Conditional Statistical Parity}}&\multirow{5}{=}{\parbox{0.1\textwidth}{Reasons for Selection}}& Important Non-protected Features& Realizing that legitimate features, like employment, savings, and job type, as important predictors for accurate decision-making, should also be included in group distribution.& They're all predicting factors.  And they are important, even if you're trying to eliminate bias. (P10)&7&7\\

    \cline{3-7}
    & & Non-protected Feature-specific Comparisons & Expressing the effectiveness of dividing people into specific groups based on shared conditions, such as same job type for fairer comparisons while reducing the risk of introducing other forms of discrimination.& I have a management job, I will care more about the people who also have a management job. […] This is fairer. Otherwise, people with different job types will be mixed together and then calculate fairness. There's another discrimination […]. You should not let other factors influence too much. (P6)&5&8\\

     \cline{3-7}
     & & Flexible Bias Detection & Expressing the preference for enabling breaking people down further into groups based on different conditions for a more flexible analysis& It's that it can include various factors. It can be more targeted and flexible. (P5)&3&4\\

     \cline{3-7}
     & & Objective & Expressing users’ belief that dividing groups based on legitimate values are objective because it can reflect the ability of people.& It's not subjective. It is objective because it's mentioned about the ability of each person, no matter what are their characteristics. (P18)&2&3\\

     \cline{2-7}
     & Reasons for Exclusion & Non-Protected Features' Susceptibility& Expressing users’ concerns about the vulnerability in accurately capturing biases, due to the potential impact of protected features on legitimate feature values&There's a significant difference between two age groups, like in credit history and employment. Age could influence this, relationships. So, dividing them under the same condition isn't fair. (P17)&1&1\\

    \hline
\end{tabular}
}
\begin{flushright}
Continued on next page.
\end{flushright}
\end{table}

\begin{table}[H]
  \textbf{Table 7 Continued from previous page}
  \resizebox{\columnwidth}{!}{%
    \begin{tabular}
	{|p{0.15\textwidth}|p{0.1\textwidth}|p{0.20\textwidth}|p{0.2\textwidth}|p{0.25\textwidth}|p{0.05\textwidth}|p{0.05\textwidth}|}
  	\hline
  	Theme & Subtheme & Code & Code Description & Example & Pts. & Freq. \\
  	\hline

     \multirow{4}{=}{\parbox{0.1\textwidth}{Consistency}}&\multirow{3}{=}{\parbox{0.1\textwidth}{Reasons for Selection}}& Comparison with Similar Individuals& Expressing a desire to compare their results with those of similar individuals to ensure fairness.& If people similar to me predictably have good credit, I should also receive a favorable prediction. It's fair. It allows you to compare yourself with people of the same level or qualifications. (P14)&6&7\\

 \cline{3-7}
     & & Inclusivity for Individuals& Expressing a preference for its potential to prevent discrimination and ensure no one is unfairly excluded.& 'Cause it prevents discrimination like, no one's getting left out sort of thing [among similar individuals]. (P13)&2&2\\

     \cline{3-7}
     & & Decision Transparency& Expressing a desire to compare their situation with others as a means to understand their own results.& I can compare to others to understand why I cannot achieve better results. (P12)&1&1\\

    \cline{2-7}
    & Reasons for Exclusion&Similarity Is Quantified& Expressing users’ concern about the inherent subjectivity in defining and measuring similarity, and the difficulty to objectively quantify.& There is no way or hard to find a standard way to analyze and compare similarity objectively. (P16)&5&8\\

    \hline
    \multirow{4}{=}{\parbox{0.1\textwidth}{Counterfactual Fairness}}&\multirow{3}{=}{\parbox{0.1\textwidth}{Reasons for Selection}}& Irrelevance Of Protected Features& Expressing a belief that factors traditionally associated with bias, such as age, gender, or foreign worker status, i.e., protected features, should not influence AI’s prediction results.& Because it disregards their age, their gender and their foreign worker status. […] Nothing should affect it at all. (P17)&5&6\\

    \cline{3-7}
    & & Individual Characteristic-based Consideration & Expressing a preference for a detailed, targeted consideration of how characteristic (i.e., protected features) would affect an individual, as opposed to broader, more generalized assessments.& The factors that are that mostly contribute to a bias […] you're looking at it on an individual basis. […] You take their individual characteristics into account. (P10) & 4& 7\\

    	\hline
    \end{tabular}
  }
  \begin{flushright}
Continued on next page.
\end{flushright}
\end{table}

\begin{table}[H]
  \textbf{Continued from previous page}
  \resizebox{\columnwidth}{!}{%
    \begin{tabular}
	{|p{0.13\textwidth}|p{0.1\textwidth}|p{0.22\textwidth}|p{0.2\textwidth}|p{0.25\textwidth}|p{0.05\textwidth}|p{0.05\textwidth}|}
  	\hline
  	Theme & Subtheme & Code & Code Description & Example & Pts. & Freq. \\
  	\hline

\cline{3-7}
    & & Objective Qualification&Expressing a preference for counterfactual fairness because it offers a more objective and convincing way to quantify individual fairness.&It is like I can quantity the similarity because the overall information for this person is still not changed, we just change the personal characteristics, [… which] is more convincing. (P5) &3&5\\

    \cline{2-7}
    & Reasons for Exclusion&Practical Ineffectiveness& Expressing users' doubts about the effectiveness of fairness evaluation, driven by the belief that altering a single feature value inherently wouldn't impact AI predictions.& It's not that useless. But It's not that helpful. Because the digital I think they are mostly the same, and however, I don't think there are so many big, big differences. (P2)&1&2\\

    \hline
    \multirow{5}{=}{\parbox{0.1\textwidth}{Equalized Odds}}&\multirow{4}{=}{\parbox{0.1\textwidth}{Reasons for Selection}}&Disparity Control for all Parties&Expressing the users' preference for a approach to control disparities between different groups. &we need to protect those vulnerable people, [...] [rated] bad [credit] people. (P14)&4&5\\

    \cline{3-7}
    & & Integrated Fairness metric & Expressing users’ preference for Equalized Odds as it integrates different fairness metrics to help detect various bias types.&Like combine them [Equal Opportunity and Predictive equality] together, you're taking into account both biases that could take place here. This makes more sense. (P10) &3&4\\

    \cline{3-7}
    & & Comprehensive Complaint Handling & Expressing the users' preference for Equalized Odds due to its wide-ranging and thorough approach to managing and resolving complaints from all customer groups.&Although my targets are not bad people, I will choose Equalized Odds, because, we have coping strategies for each kind of customer.(P16) & 2&2\\

    \cline{3-7}
    & & Ground-Truth Label Importance & Expressing user's preference for metrics that rely on rated credits and belief that such alignment ensures a more trustworthy assessment of fairness.& I think the rated credits will be more believable part to evaluate this because. I think it's mostly come from the rated one.[…] I really think the rated credit is more important than the condition features. (P2)&1&2\\

    	\hline
    \end{tabular}
  }
  \begin{flushright}
Continued on next page.
\end{flushright}
\end{table}

\begin{table}[H]
  \textbf{Continued from previous page}
  \resizebox{\columnwidth}{!}{%
    \begin{tabular}
	{|p{0.13\textwidth}|p{0.1\textwidth}|p{0.22\textwidth}|p{0.2\textwidth}|p{0.25\textwidth}|p{0.05\textwidth}|p{0.05\textwidth}|}
  	\hline
  	Theme & Subtheme & Code & Code Description & Example & Pts. & Freq. \\
  	\hline

 \cline{2-7}
    & Reasons for Exclusion&Redundancy&Expressing that using Equalized Odds introduces a redundancy of fairness considerations, when users think pursuing fairness should be sufficient for qualified people.&I think it's good just to focus on the good part. (P4)&1&1\\

    \hline
    \multirow{2}{=}{\parbox{0.1\textwidth}{Equal Opportunity}}&Reasons for Selection&Qualified People Focus Only&Expressing a preference for only focusing on individuals with good qualifications when making decisions&I think in the loan applications, we have to care about if the customer has good qualifications. If the customer has already been judged as bad credit by humans, we do not need to care about these sections at all.  (P4)&5&11\\

    \cline{2-7}
    & Reasons for Exclusion&Incompleteness&Expressing that Equal Opportunity only assesses fairness among qualified people, overlooking comprehensive fairness when users prefer fairness for all, regardless of qualifications.&I think all the people are important. The overall, the whole. (P6)&1&1\\

    \hline
    \multirow{2}{=}{\parbox{0.1\textwidth}{Predictive Equality}}&Reasons for Selection&Risk Considerations on Non-Qualified People&Expressing a perspective focused on risk management and ethical considerations which suggests a belief in the need for both accurate and fair decisions.&Like bad people should not be provided with loans. […]. We should provide more care in making sure that bad people don't get loads. (P15)&2&4\\

    \cline{2-7}
    & Reasons for Exclusion&Unnecessary to consider non-qualified individuals&Expressing that using Predictive Equality considering fairness among non-qualified people is unnecessary and meaningless, when users think pursuing fairness is only for qualified people.&I will not consider the fairness among bad people. Meaningless. For the loan application, […] the final purpose is to provide a better service to those qualified, potential customers. (P16)&4&6\\

    	\hline
    \end{tabular}
  }
  \begin{flushright}
Continued on next page.
\end{flushright}
\end{table}

\begin{table}[H]
  \textbf{Continued from previous page}
  \resizebox{\columnwidth}{!}{%
    \begin{tabular}
	{|p{0.13\textwidth}|p{0.1\textwidth}|p{0.22\textwidth}|p{0.2\textwidth}|p{0.25\textwidth}|p{0.05\textwidth}|p{0.05\textwidth}|}
  	\hline
  	Theme & Subtheme & Code & Code Description & Example & Pts. & Freq. \\
  	\hline

 \hline
    \multirow{2}{=}{\parbox{0.1\textwidth}{Outcome Test}}&Reasons for Selection&Direct Fairness Assessment&Expressing the users' belief that the fairness of a model can be assessed by 
directly comparing the model's predictions with actual labels.&With good predictions and bad predictions, we are trying to figure out which are the good which are the bad. And based on that, we are trying to […] get [fair] predictions. I feel like that's the case. P(18)&3&4\\

    \cline{2-7}
    & Reasons for Exclusion&Over-Reliance on AI's Performance&Expressing unease with first filtering out the AI's 'good' predictions and then assessing the proportion of real good ones, especially when doubting AI's effectiveness.&Because the outcome test is pretty rely on the AI’s predictions. our current model accuracy is only 76\%. (P9)&2&2\\

    \hline
    \multirow{2}{=}{\parbox{0.1\textwidth}{Demographic Parity}}&Reasons for Selection&Holistic Fairness Assessment&Expressing a preference for considering a holistic assessment where 
the total population of each group is accounted for, rather than focusing on further refined groups based on types of ground-truth labels or prediction results.
    &The number of people you judged or cared about is more comprehensive. If you just focused on like rated good or rated bad […] you always divided them into two. But demographic parity could give me an overall percentage in all group people. (P5)&2&2\\

    \cline{2-7}
    & Reasons for Exclusion&Neglect Individual Qualifications&Expressing users’ concerns about the neglect of people’s qualifications in the pursuit of fairness.&It doesn't actually care about real good or bad. It just gives importance to the gender equality thing. (P15)&5&7\\

    	\hline
    \end{tabular}
  }
\end{table}
\clearpage

\subsubsection{Participants' Feedback on UI Design}
\label{app:Participants' Feedback on UI Design}
Below is the codebook we generated while analyzing participants feedback.

\begin{table}[H]
  \caption{Codebook-Participants' Feedback on UI Design}
  \label{tab:Codebook-UIDesign}
  \captionsetup{skip=0pt} 
  \resizebox{\columnwidth}{!}{%
    \begin{tabular}
	{|p{0.1\textwidth}|p{0.25\textwidth}|p{0.25\textwidth}|p{0.3\textwidth}|p{0.05\textwidth}|p{0.05\textwidth}|}
  	\hline
  	Subtheme & Code & Code Description & Example & Pts. & Freq. \\
  	\hline

    \multirow{5}{=}{\parbox{0.1\textwidth}{Positive Feedback}}& AI Concepts and Model Explanations& Expressing that effective model explanation helps lay users easily grasp fundamental AI concepts, Model Performance and application scenarios.& I think model explanation module introducing definitions, how to use. I think before using it, model explanation this module let me know how to use, get an overall picture. (P3)&5&5\\

    \cline{2-6}
     &Intuitive Visualization for Fairness Metric Explanations& Expressing that the transformation of intricate metric formulas into user-friendly visual formats (e.g., colour coding and visual elements) effectively help with metric understanding. & I have no AI expertise and I don’t need math or very little math for my major. I am not good at it. So I just need to replace the colors with the titles [in metric formulas] […] With a glance, I can see how the value is calculated. (P3)& 6&6\\

     \cline{2-6}
     &Interactive Exploration &Expressing the positive role of interactive elements (e.g., clicking on images or hovering on graphs) in data visualization.  & I can click each image, impressive and useful. I can check each individual. […] The difference bar shows the difference value between these two groups and the image shows how this difference value is calculated. (P3)&2&2\\

    \cline{2-6}
     & Real-Time Updates to Fairness Results& Expressing the positive role of real-time interactive label modification in tracking instant fairness changes to help users make top three choices, and facilitate re-thinking and double-checking top3 preferences.& After I changing the labels, I can check the changes. I think it is necessary. […] After I chose my top 3 fairness metrics, I think using this function is like a process of reflection. To reflect or re-think how the fairness results changed. (P14) &8&14\\

     \cline{2-6}
    & Functionality&Expressing positive feedback on comprehensive UI functionality. & It's quite an elegant piece of technology you've got there and you think this you know that all this is it's helpful. (P1)&5&10\\

    \cline{2-6}
     &Aesthetics & Expressing positive feedback on elegant, visually appealing layout in overall UI design.& This is a prototype I'd imagine it's like is the first version for time. Yeah. but I'd say it's it's good. It's really good so far. (P13)&5&6\\

    \hline
\end{tabular}
}
\begin{flushright}
Continued on next page.
\end{flushright}
\end{table}

\begin{table}[H]
  \textbf{Table 8 Continued from previous page}
  \resizebox{\columnwidth}{!}{%
    \begin{tabular}
	{|p{0.12\textwidth}|p{0.25\textwidth}|p{0.25\textwidth}|p{0.3\textwidth}|p{0.03\textwidth}|p{0.05\textwidth}|}
  	\hline
  	Subtheme & Code & Code Description & Example & Pts. & Freq. \\
  	\hline

\hline
    \multirow{4}{=}{\parbox{0.1\textwidth}{Suggestions for Improving UI Design}}& Model Transparency& Expressing that users engage with AI predictions, they seek to identify the specific factors that have a greater impact on the final results.& But the person reviewing them might also want to look at the reason for that. Maybe you could […] give me what feature was it about the customer that caused the decision to be bad. (P13)&2&3\\

    \cline{2-6}
    & Practical Implications of Fairness Metrics&Expressing that users' need for both technical knowledge of metric calculations and insights into their practical applications and real-world implications, addressing difficulties in metric interpretation due to abstract naming and a lack of contextual clarity. & If each metric combining with a context explanation would be better for me to select. (P16) I can understand how they are calculated but don't know what the point is. (P3)&6&10\\

   \cline{2-6}
     &Simplified Data Visualization \& Explanation &Expressing the need to tailor data visualization formats to user expertise, starting with basic, universally accessible formats like bar and pie charts for conveying fairness concepts to lay users. & Maybe also like bar charts are just easier to read for everybody. So expressed in more like pie chart formats or easier formats or like regular bar graphs, that might be easier for like fully lay people to understand if that makes sense. (P10) &3&3\\

     \cline{2-6}
    & Simplifying  Information Quantity& Expressing the participant's viewpoint on limiting the number of data points visible to users at any given time, aiming to alleviate cognitive overload and prevent confusion that might arise from an abundance of data.& At least at least for the UI initially like maybe have less data points visible […] why is there so much data? What am I supposed to get gather from that? […] let's say it might be more difficult for them to gather anything. (P10)&3&4\\

    \hline
   \multirow{2}{=}{\parbox{0.1\textwidth}{User Experience}}  &Cognitively Challenging &Expressing a user experience that requires a high level of mental effort and concentration. & I got a little bit lost at first so until I finish the whole process. […] It makes a lot of time to understand. Get to half an hour for me. (P8)&8&12\\

   \cline{2-6}
     &Engaging   &Expressing users find this UI prototype and the AI fairness content engaging and interesting. & I think it is very interesting for me. I can understand your research. (P6)&10&10\\

    \hline
\end{tabular}
}
\end{table}

\end{document}